%% file: aiwild.tex
\PassOptionsToPackage{numbers,square,sort&compress}{natbib}

\documentclass{article}
\usepackage[sglblindworkshop,final]{neurips_2026}
\workshoptitle{AIWILD}   % OPEN: full official workshop name to confirm
\newif\ifanon \anonfalse
\newif\ifwmac \wmacfalse
\newif\ifaiwild \aiwildtrue
\usepackage[utf8]{inputenc}
\usepackage[T1]{fontenc}
\usepackage{microtype}
\usepackage{graphicx}
\usepackage{url}
\usepackage{hyperref}
\hypersetup{colorlinks=true,allcolors=blue!50!black}
\input{macros}

\title{Do Multi-Agents Dream of Electric Screens?\\Achieving 100\% on AndroidWorld Through Task Decomposition}
\author{%
  Pierre-Louis Favreau$^{1}$ \quad Jean-Pierre Lo$^{1}$ \quad Clement Guiguet$^{1}$ \quad Charles Simon-Meunier$^{1}$ \\
  \textbf{Nicolas Dehandschoewercker}$^{1}$ \quad \textbf{Allen G. Roush}$^{2}$ \\
  \textbf{Judah Goldfeder}$^{3}$ \quad \textbf{Ravid Shwartz-Ziv}$^{4}$ \\
  $^{1}$Minitap \quad $^{2}$Thoughtworks \quad $^{3}$Columbia University \quad $^{4}$New York University \\
  \texttt{nico@minitap.ai}
}

\begin{document}
\maketitle
\begin{abstract}
\input{sections/abstract}
\end{abstract}

\input{sections/intro}

\input{sections/problem}

\input{sections/architecture}

\input{sections/experiments}

\input{sections/reliability}
\input{sections/conclusion}

\bibliographystyle{plainnat}
\bibliography{references}

% Impact statement (not required for NeurIPS workshops) placed after the references.
\input{sections/impact}

\newpage
\appendix
\input{sections/appendix}
\end{document}

%% file: macros.tex
\usepackage{xcolor}
\usepackage{booktabs}
\usepackage{amsmath}
\usepackage{amssymb}
\usepackage{mathtools}
\usepackage{tikz}
\usetikzlibrary{arrows.meta,fit,backgrounds,shapes.geometric}
\usepackage{tcolorbox}

\newcommand{\TBD}[1]{\textcolor{red}{\textbf{[TBD:~#1]}}}
\newcommand{\fitcol}[1]{\resizebox{\ifdim\width>\linewidth\linewidth\else\width\fi}{!}{#1}}

\newif\ifversionB \versionBtrue
\newif\ifabstractB \abstractBfalse   % version-B abstract sentence: prepared, switched on only after the finals (Pilou's call)
\newif\ifbmoved \ifaiwild\ifversionB\bmovedtrue\fi\fi   % AIWILD version-B build: part of Section 4.1 moves to Appendix C (page budget)

\ifanon
  \newcommand{\sys}{\textsc{MobileMA}}
  \newcommand{\sysfull}{\textsc{MobileMA}}
  \newcommand{\repo}{[code link withheld for review]}
  \newcommand{\codepath}{agent package of the released code}
  
  \newcommand{\agentsha}{commit withheld for review}
  \newcommand{\finalref}{at a commit withheld for review}
  \newcommand{\cloudfirst}{cloud-hosted Android emulators}
  \newcommand{\farm}{a physical device farm}
\else
  \newcommand{\sys}{Minitap}
  \newcommand{\sysfull}{Minitap's mobile-use}
  \newcommand{\repo}{\url{https://github.com/minitap-ai/mobile-use}}
  \newcommand{\codepath}{\texttt{minitap/mobile\_use/agents/}, v3.4.0}
  
  \newcommand{\agentsha}{\texttt{988b436}}
  \newcommand{\finalref}{at commit \resBFreezePin{}, tagged \texttt{androidworld-paper-2026} (\url{https://github.com/minitap-ai/mobile-use/tree/androidworld-paper-2026})}
  \newcommand{\cloudfirst}{Android emulators in Minitap's cloud device infrastructure}
  \newcommand{\farm}{our device farm}
\fi

\newtcolorbox{takeaway}{
  colback=blue!5!white,
  colframe=blue!50!black,
  boxrule=0.5pt,
  arc=2pt,
  left=4pt,
  right=4pt,
  top=2pt,
  bottom=2pt,
  fontupper=\small
}

\input{results/autogen}

\providecommand{\nPhys}{\TBD{P1: n}}
\providecommand{\nEmu}{\TBD{P1: n}}
\providecommand{\resPhysHeadlineSRtext}{\TBD{P3: SR}}
\providecommand{\resEmuHeadlineSRbrief}{\resEmuSRFull\,\%}
\providecommand{\resBLimitSteps}{\TBD{P2: B graph-step limit}}
\providecommand{\resBLimitSeconds}{\TBD{P2: B wall-clock cap (s)}}
\providecommand{\resBFirstAfterText}{\TBD{P2 final: FirstEventAfterStartTime regression sentence}}
\providecommand{\resBMiniVerdictText}{\TBD{P2 MiniWoB check: verdict sentence}}
\providecommand{\resBInfraLimitedText}{\TBD{P2 final: infra-limited statement}}
\providecommand{\resBFreezeText}{\TBD{P2 freeze: criterion, gate outcome, best-evidenced rule if used}}
\ifanon\renewcommand{\resBFreezePin}{a commit withheld for review}\fi   % WMAC: no commit hash (anonymity)
\providecommand{\resBMiniDeltaCIrange}{\TBD{P2 MiniWoB check: CI bounds}}
\providecommand{\resBMiniDeltaPts}{\TBD{P2 MiniWoB check: unsigned delta}}
\providecommand{\resBRouteRiskText}{\TBD{P2: route residual risk, e.g. about 1 in 8 runs}}

\providecommand{\resBOsmTrackText}{\TBD{P2: OsmAndTrack withdrawn readiness attempt, one sentence}}
\providecommand{\resOssFallbackShare}{\TBD{P2: share}}
\providecommand{\resServedOutsideText}{\TBD{P2: calls served outside the allowed set}}
\providecommand{\resAwDecompSig}{U}
\newcommand{\awdecompS}{S}\newcommand{\awdecompN}{N}
\newcommand{\awdecomp}[2]{\ifx\resAwDecompSig\awdecompS #1\else\ifx\resAwDecompSig\awdecompN #2\else\TBD{if the Cloud-90 decomposition effect is significant:}~\textcolor{red}{#1}~\TBD{otherwise:}~\textcolor{red}{#2}\fi\fi}
\providecommand{\resMiniNonIntRewardText}{\TBD{P4: D5 cells per configuration, final runs}}
\providecommand{\resMiniDestroyedText}{\TBD{P4: D7 cells per configuration, final runs}}
\providecommand{\resMiniInfraLostText}{\TBD{P4: MiniWoB cells lost to infrastructure, final runs}}

%% file: results/autogen.tex
\providecommand{\TBD}[1]{\textcolor{red}{\textbf{[TBD:~#1]}}}

\newcommand{\resHeadlineSRtext}{83.2\,\% (95\,\% CI 76.7--89.2)}

\newcommand{\resHeadlinePassText}{pass@2 = 90.5\,\%, pass\^{}2 = 75.9\,\%}

\newcommand{\resHeadlineBrittleText}{17 of 116 tasks (15\,\%) are solved in some but not all of the 2 runs: pass@2 90.5\,\%, mean 83.2\,\%, pass\^{}2 75.9\,\%}

\newcommand{\resTimeoutRate}{0.4\,\% (1 of 232 task runs)}
\newcommand{\resTimeToSuccess}{median 4.1 min (IQR 2.6--8.3, p90 12.3, max 24.9; $n=193$)}

\newcommand{\resLoopTriggers}{0.19 per task on average (43 triggers over 232 task runs; 22 runs with at least one)}
\newcommand{\resTextCheck}{387 read-back checks in 111 task runs (345 MATCH, 14 CONTAINS, 0 CASE\_ONLY, 14 MISMATCH, 14 UNVERIFIABLE); mismatch rate 3.8\,\%; 2 of 14 mismatches followed by a matching re-entry}
\newcommand{\resVideoErrText}{the video analyzer's provider rejected 15 recordings as malformed (HTTP 400) in 3 of 232 task runs; 2 failed task runs had no successful video analysis (MarkorTranscribeReceipt (seed 30), MarkorTranscribeVideo (seed 31))}

\newcommand{\resPriceDate}{2026-11-12}
\newcommand{\resTabRuns}{{\small\begin{tabular}{@{}lrlrrr@{}}\toprule Config & Seed & Date & Solved & SR (\%) & Infra re-queued \\ \midrule Full system & 30 & 2026-11-10 & 91/116 & 78.4 & 5 \\ Full system & 31 & 2026-11-10 & 102/116 & 87.9 & 5 \\ Single-agent baseline & 30 & 2026-11-10 & 51/116 & 44.0 & 3 \\ Single-agent baseline & 31 & 2026-11-10 & 51/116 & 44.0 & 3 \\ \bottomrule\end{tabular}}}
\newcommand{\resTabFlip}{{\small\begin{tabular}{@{}lcc@{}}\toprule Task & seed 30 & seed 31 \\ \midrule AudioRecorderRecordAudioWithFileName & $\times$ & \checkmark \\ BrowserMultiply & $\times$ & \checkmark \\ ExpenseAddMultiple & $\times$ & \checkmark \\ ExpenseAddMultipleFromGallery & $\times$ & \checkmark \\ ExpenseDeleteDuplicates & $\times$ & \checkmark \\ ExpenseDeleteMultiple2 & $\times$ & \checkmark \\ FilesMoveFile & $\times$ & \checkmark \\ MarkorTranscribeReceipt & $\times$ & \checkmark \\ MarkorTranscribeVideo & \checkmark & $\times$ \\ NotesTodoItemCount & $\times$ & \checkmark \\ OsmAndTrack & $\times$ & \checkmark \\ RecipeAddMultipleRecipesFromMarkor2 & $\times$ & \checkmark \\ RetroPlayingQueue & \checkmark & $\times$ \\ RetroPlaylistDuration & \checkmark & $\times$ \\ SportsTrackerTotalDistanceForCategoryOverInterval & $\times$ & \checkmark \\ SystemBrightnessMax & $\times$ & \checkmark \\ TasksDueOnDate & $\times$ & \checkmark \\ \bottomrule\end{tabular}}}
\newcommand{\resTabPrices}{{\small\begin{tabular}{@{}lrrr@{}}\toprule Model & \$/M in & \$/M out & Calls \\ \midrule \texttt{google/gemini-3-flash-preview} & 0.500 & 3.000 & 19 \\ \texttt{google/gemini-3.1-pro-preview} & 2.000 & 12.000 & 50234 \\ \texttt{meta-llama/llama-3.3-70b-instruct} & 0.710 & 0.710 & 30514 \\ \texttt{meta-llama/llama-4-scout} & 0.107 & 0.232 & 2245 \\ \texttt{openai/gpt-5-nano} & 0.050 & 0.400 & 3395 \\ \texttt{openai/gpt-oss-120b} & 0.050 & 0.250 & 8026 \\ \texttt{qwen/qwen3-vl-8b-instruct} & 0.250 & 0.750 & 9920 \\ \bottomrule\end{tabular}}\\{\scriptsize Prices fitted to the logged per-call costs, calls up to 2026-11-12.}}
\newcommand{\resFigTTS}{\includegraphics[width=\columnwidth]{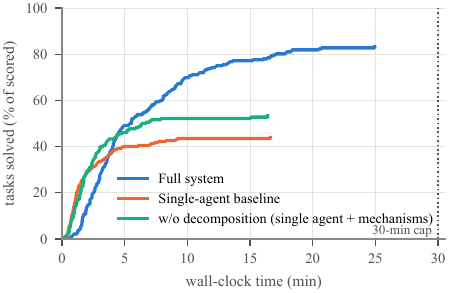}}
\newcommand{\resFigRoles}{\includegraphics[width=\textwidth]{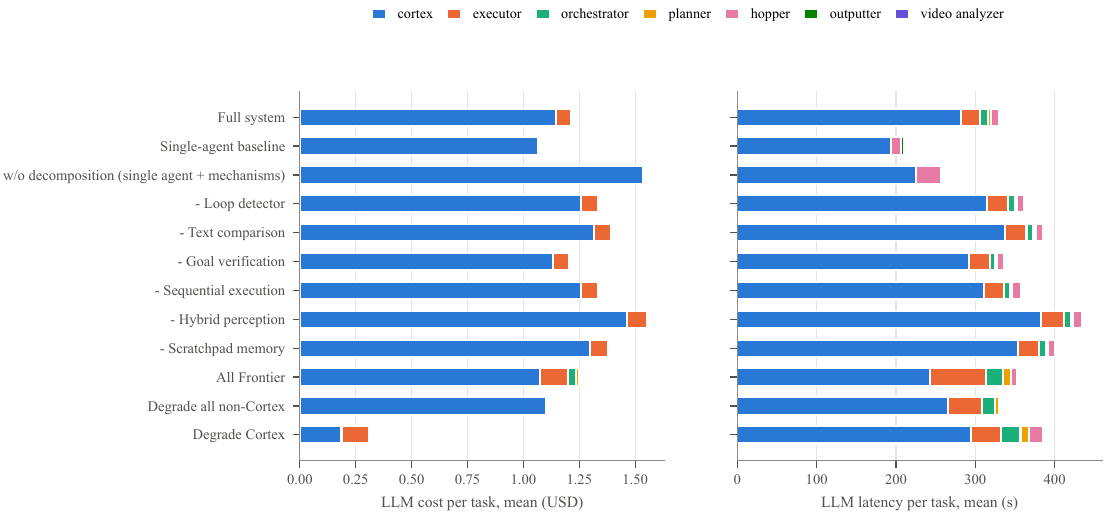}}
\newcommand{\resFigHeatmap}{\includegraphics[width=\columnwidth]{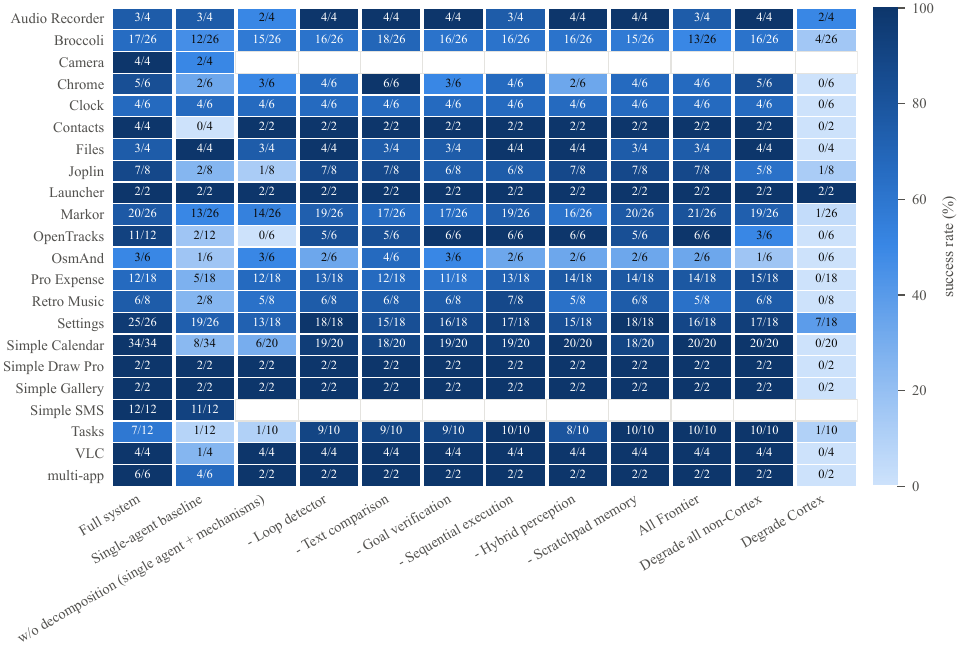}}

\newcommand{\resPhysHeadlineSRtext}{93.8\,\% (95\,\% CI 81.2--100.0)}

\newcommand{\resEmuHeadlineSRbrief}{100.0\,\% (95\,\% CI 72.2--100.0)}  % Wilson on tasks; bootstrap degenerate

\newcommand{\resSRcellFull}{79.4 {\scriptsize[71.1, 87.2]}}

\newcommand{\resCostFull}{1.32}

\newcommand{\resCallsFull}{26.0}

\newcommand{\resSRcellSingleAgent}{41.7 {\scriptsize[32.2, 51.7]}}

\newcommand{\resSRcellDecompOnly}{53.3 {\scriptsize[42.8, 63.3]}}

\newcommand{\resCostDecompOnly}{1.54}

\newcommand{\resSRcellNoLoop}{80.0 {\scriptsize[72.2, 87.8]}}

\newcommand{\resSRcellNoTextCheck}{78.9 {\scriptsize[71.1, 86.7]}}

\newcommand{\resSRcellNoGoalVerif}{76.1 {\scriptsize[67.8, 83.9]}}

\newcommand{\resSRcellNoSequential}{80.0 {\scriptsize[71.7, 87.8]}}

\newcommand{\resSRcellNoHybrid}{76.1 {\scriptsize[67.2, 84.4]}}

\newcommand{\resSRcellNoScratchpad}{80.0 {\scriptsize[71.7, 87.8]}}

\newcommand{\resSRcellAllFrontier}{78.9 {\scriptsize[70.6, 86.7]}}

\newcommand{\resCostAllFrontier}{1.26}

\newcommand{\resSRcellNonCortexBudget}{79.4 {\scriptsize[71.1, 87.2]}}

\newcommand{\resCostNonCortexBudget}{1.12}

\newcommand{\resCallsNonCortexBudget}{23.5}

\newcommand{\resSRcellCortexBudget}{10.0 {\scriptsize[4.4, 16.1]}}
\newcommand{\resSRtextCortexBudget}{10.0\,\% (95\,\% CI 4.4--16.1)}

\newcommand{\resCostCortexBudget}{0.32}

\newcommand{\resCallsCortexBudget}{77.8}

\newcommand{\resEffDecompositionCI}{${+}26.1$ [${+}16.7$, ${+}35.6$]}

\newcommand{\resEffMechanismsCI}{${+}11.7$ [${+}2.2$, ${+}21.1$]}

\newcommand{\resDeltaCISingleAgent}{$-37.8$ [$-47.8$, $-27.8$]}
\newcommand{\resDeltacellSingleAgent}{$-37.8$ {\scriptsize[$-47.8$, $-27.8$]}}

\newcommand{\resDeltaCIDecompOnly}{$-26.1$ [$-35.6$, $-16.7$]}
\newcommand{\resDeltacellDecompOnly}{$-26.1$ {\scriptsize[$-35.6$, $-16.7$]}}

\newcommand{\resDeltaCINoLoop}{${+}0.6$ [$-5.0$, ${+}6.1$]}
\newcommand{\resDeltacellNoLoop}{${+}0.6$ {\scriptsize[$-5.0$, ${+}6.1$]}}

\newcommand{\resDeltaCINoTextCheck}{$-0.6$ [$-7.2$, ${+}6.7$]}
\newcommand{\resDeltacellNoTextCheck}{$-0.6$ {\scriptsize[$-7.2$, ${+}6.7$]}}

\newcommand{\resDeltacellNoGoalVerif}{$-3.3$ {\scriptsize[$-10.6$, ${+}3.3$]}}

\newcommand{\resDeltacellNoSequential}{${+}0.6$ {\scriptsize[$-5.0$, ${+}6.7$]}}

\newcommand{\resDeltacellNoHybrid}{$-3.3$ {\scriptsize[$-10.6$, ${+}3.3$]}}

\newcommand{\resDeltacellNoScratchpad}{${+}0.6$ {\scriptsize[$-5.6$, ${+}7.2$]}}

\newcommand{\resRDeltaCIFull}{${+}0.6$ [$-6.1$, ${+}7.2$]}
\newcommand{\resRDeltacellFull}{${+}0.6$ {\scriptsize[$-6.1$, ${+}7.2$]}}

\newcommand{\resRDeltaNonCortexBudget}{${+}0.6$}
\newcommand{\resRDeltaCINonCortexBudget}{${+}0.6$ [$-6.1$, ${+}7.2$]}
\newcommand{\resRDeltacellNonCortexBudget}{${+}0.6$ {\scriptsize[$-6.1$, ${+}7.2$]}}

\newcommand{\resRDeltacellCortexBudget}{$-68.9$ {\scriptsize[$-77.8$, $-60.0$]}}

\newcommand{\resPDDeltaCortexBudget}{$-69.4$}
\newcommand{\resPDDeltaCICortexBudget}{$-69.4$ [$-77.8$, $-60.6$]}

\newcommand{\resLimHeadlineSRtext}{79.4\,\% (95\,\% CI 71.1--87.2)}

\newcommand{\resMiniSRcellFull}{70.2 {\scriptsize[58.9, 80.6]}}

\newcommand{\resMiniCostFull}{0.63}

\newcommand{\resMiniSRcellSingleAgent}{50.0 {\scriptsize[37.9, 62.1]}}
\newcommand{\resMiniSRtextSingleAgent}{50.0\,\% (95\,\% CI 37.9--62.1)}

\newcommand{\resMiniSRcellDecompOnly}{78.2 {\scriptsize[67.7, 87.1]}}
\newcommand{\resMiniSRtextDecompOnly}{78.2\,\% (95\,\% CI 67.7--87.1)}

\newcommand{\resMiniCostDecompOnly}{0.38}

\newcommand{\resMiniEffDecompositionCI}{$-8.1$ [$-18.5$, ${+}2.4$]}

\newcommand{\resMiniEffDecompositionP}{0.076}

\newcommand{\resMiniEffMechanismsCI}{${+}28.2$ [${+}17.7$, ${+}39.5$]}

\newcommand{\resMiniHeadlineSRtext}{70.2\,\% (95\,\% CI 58.9--80.6)}

\newcommand{\resMiniHeadlinePassText}{pass@2 = 80.6\,\%, pass\^{}2 = 59.7\,\%}

\newcommand{\resMiniAllSRcellFull}{64.1 {\scriptsize[54.9, 73.4]}}

\newcommand{\resMiniAllCostFull}{0.54}

\newcommand{\resMiniAllWallMedFull}{96}

\newcommand{\resMiniAllSRcellSingleAgent}{46.7 {\scriptsize[37.0, 56.5]}}

\newcommand{\resMiniAllCostSingleAgent}{0.20}

\newcommand{\resMiniAllWallMedSingleAgent}{26}

\newcommand{\resMiniAllSRcellDecompOnly}{70.7 {\scriptsize[61.4, 79.3]}}

\newcommand{\resMiniAllCostDecompOnly}{0.34}

\newcommand{\resMiniAllWallMedDecompOnly}{38}

\newcommand{\resMiniAllHeadlineSRtext}{64.1\,\% (95\,\% CI 54.9--73.4)}

\newcommand{\resWallHeadVsSingle}{281\,s (IQR 166--571) for the full system vs. 137\,s (IQR 73--300) for the single-agent baseline}
\newcommand{\resToolLatencyMs}{537\,ms median per action}
\newcommand{\resSingleAgentSRtext}{41.7\,\% (95\,\% CI 32.2--51.7)}
\newcommand{\resFigPareto}{\includegraphics[width=\columnwidth]{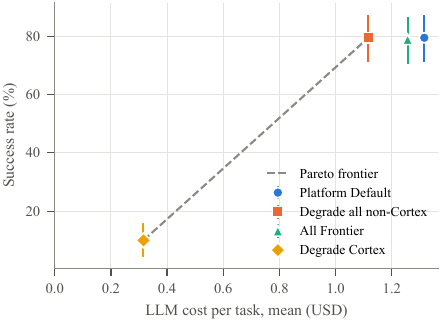}}
\newcommand{\resFigAblation}{\includegraphics[width=\columnwidth]{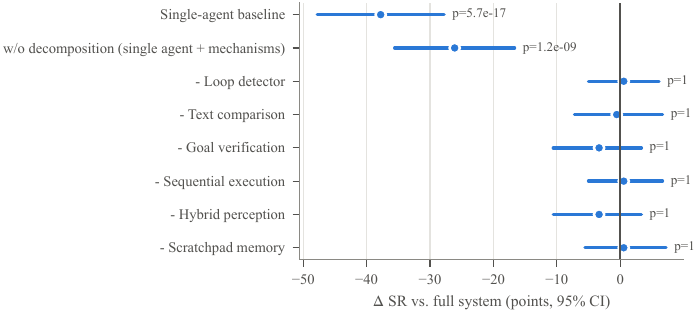}}

\newcommand{\resOvhCells}{40}
\newcommand{\resOvhTasks}{10}
\newcommand{\resOvhDwall}{${+}2.3$}
\newcommand{\resOvhDstep}{$-5.0$}
\newcommand{\resOvhDwallRange}{$-64.6$ to ${+}33.9$}
\newcommand{\resOvhTap}{${+}2.5$}
\newcommand{\resOvhInputText}{${+}31.3$}
\newcommand{\resOvhInputTextN}{13}
\newcommand{\resOvhAgree}{8 of 10}
\newcommand{\resOvhCriteriaMet}{met}
\newcommand{\resOvhVideos}{20}
\newcommand{\resOvhVideoMB}{2.94}
\newcommand{\resOvhVideoMBRange}{0.02--72.94}

\newcommand{\resOsmWipedText}{about 55 of 72}  % approximate: reconstructed from device and run order
\newcommand{\resContactSetupN}{8}

\newcommand{\resBUnscoredN}{0}
\newcommand{\resBUnscoredShare}{0.0}

\newcommand{\resRecipeIngredientsN}{120}
\newcommand{\resRecipeCapsN}{8}
\newcommand{\resRecipeACells}{24}
\newcommand{\resRecipeAAffected}{0}

\newcommand{\resRecipeFinalText}{none of the final instances is affected}
\newcommand{\resCalNowACells}{8}
\newcommand{\resCalNowAAffected}{0}

\newcommand{\resCalNowFinalText}{none of the final instances is affected}
\newcommand{\resSportsACells}{60}
\newcommand{\resSportsAAffected}{0}

\newcommand{\resSportsFinalText}{1 of the 15 final instances is affected (SportsTrackerTotalDurationForCategoryThisWeek with seed 72)}
\newcommand{\resSportsFinalAmbigText}{one final instance, SportsTrackerTotalDurationForCategoryThisWeek with seed 72, has an ambiguous goal}
\newcommand{\resFirstAfterACells}{24}
\newcommand{\resFirstAfterASolved}{16}

\newcommand{\resBAWSRtext}{98.0\,\% (95\,\% CI 96.0--99.7)}
\newcommand{\resBAWPassAtK}{100.0}
\newcommand{\resBAWPassAtKtext}{100.0\,\% (95\,\% CI 96.8--100.0)}

\newcommand{\resBAWPassHatKtext}{94.0\,\% (95\,\% CI 89.7--97.4)}
\newcommand{\resBAWNeverSolved}{none}

\newcommand{\resBCloudSRtext}{99.2\,\% (95\,\% CI 97.7--100.0)}
\newcommand{\resBCloudPassAtK}{100.0}
\newcommand{\resBCloudPassAtKtext}{100.0\,\% (95\,\% CI 95.8--100.0)}

\newcommand{\resBCloudPassHatKtext}{97.7\,\% (95\,\% CI 94.3--100.0)}
\newcommand{\resBCloudNeverSolved}{none}

\newcommand{\resBPhysSRtext}{93.8\,\% (95\,\% CI 83.3--100.0)}
\newcommand{\resBPhysPassAtK}{100.0}
\newcommand{\resBPhysPassAtKtext}{100.0\,\% (95\,\% CI 80.6--100.0)}

\newcommand{\resBPhysPassHatKtext}{81.2\,\% (95\,\% CI 62.5--100.0)}
\newcommand{\resBPhysNeverSolved}{none}

\newcommand{\resBEmuSRtext}{94.9\,\% (95\,\% CI 84.6--100.0)}
\newcommand{\resBEmuPassAtK}{100.0}
\newcommand{\resBEmuPassAtKtext}{100.0\,\% (95\,\% CI 77.2--100.0)}

\newcommand{\resBEmuPassHatKtext}{84.6\,\% (95\,\% CI 61.5--100.0)}
\newcommand{\resBEmuNeverSolved}{none}

\newcommand{\resBFreshSRtext}{97.5\,\% (95\,\% CI 94.9--99.6)}

\newcommand{\resBFreshPassAtKtext}{100.0\,\% (95\,\% CI 96.0--100.0)}

\newcommand{\resBFreshPassHatKtext}{92.4\,\% (95\,\% CI 87.0--97.8)}
\newcommand{\resBFreshNeverSolved}{none}
\newcommand{\resBK}{3}
\newcommand{\resBNIterations}{27}
\newcommand{\resBNFreezeChecks}{4}
\newcommand{\resBFreezePin}{\texttt{10908eb}}
\newcommand{\resBMiniDeltaCI}{{+}7.1 [-0.5, {+}14.7]}
\newcommand{\resBMiniDeltaP}{0.035}

\newcommand{\resBMiniSubsetDeltaCI}{{+}8.1 [-1.6, {+}17.7]}
\newcommand{\resBMiniDeltaCIrange}{$-0.5$ to ${+}14.7$}
\newcommand{\resBMiniDeltaPts}{7.1}
\newcommand{\resBLimitSteps}{1500}
\newcommand{\resBLimitSeconds}{3600}
\newcommand{\resBFinalSeeds}{71, 72 and 73}
\newcommand{\resBDevSeeds}{30, 31 and 40--49}
\newcommand{\resBNCloud}{87}
\newcommand{\resBNPhys}{16}
\newcommand{\resBNEmu}{13}
\newcommand{\resBRequeueRounds}{2}
\newcommand{\resBFirstAfterText}{SimpleCalendarFirstEventAfterStartTime, solved in 2 of 2 runs of the initial system, is solved in 3 of 3 runs of the final system (the task is solvable only under an inclusive reading of ``after''; Appendix~\ref{app:protocol}).}
\newcommand{\resBMiniVerdictText}{no regression (the 95\,\% CI lower bound is above $-5$ points)}
\newcommand{\resBInfraLimitedText}{the final evaluation is not infra-limited}
\newcommand{\resBFreezeText}{The frozen pin \resBFreezePin{} was chosen by a pre-declared targeted procedure: every task changed since the best-evidenced development pin passed a fixed per-task bar on fresh held-out seeds (5 of 5 for tasks that had failed, 3 of 3 otherwise), and every other task kept its earlier evidence; no pin met the earlier freeze gates, and no full check was run on the frozen pin.}
\newcommand{\resBAWBrittleText}{7 of 116 tasks (6\,\%) are solved in some but not all of the 3 runs: pass@3 100.0\,\%, mean 98.0\,\%, pass\^{}3 94.0\,\%}
\newcommand{\resBAWFlipN}{7}
\newcommand{\resBTabFlip}{{\small\begin{tabular}{@{}lccc@{}}\toprule Task & seed 71 & seed 72 & seed 73 \\ \midrule MarkorCreateNoteAndSms & \checkmark & $\times$ & \checkmark \\ NotesTodoItemCount & $\times$ & \checkmark & \checkmark \\ RecipeDeleteMultipleRecipesWithNoise & \checkmark & \checkmark & $\times$ \\ SimpleSmsSendReceivedAddress & $\times$ & \checkmark & \checkmark \\ SportsTrackerActivitiesCountForWeek & \checkmark & $\times$ & \checkmark \\ SportsTrackerLongestDistanceActivity & \checkmark & $\times$ & \checkmark \\ SportsTrackerTotalDurationForCategoryThisWeek & \checkmark & $\times$ & \checkmark \\ \bottomrule\end{tabular}}}

\newcommand{\resBRouteRiskText}{1 of 13 scored runs on the frozen pin, including the withdrawn attempts}
\newcommand{\resBOsmTrackText}{Two readiness attempts on OsmAndTrack were withdrawn for operator interference before their results were read; one scored 0, by the route-procedure failure listed under known limitations, and none is a counted row.}
\newcommand{\resExpMarkerACells}{24}
\newcommand{\resExpMarkerASolved}{0}
\newcommand{\resStepsAExceedShare}{18.1}
\newcommand{\resStepsAExceedText}{35 of 193}
\newcommand{\resStepsAMedian}{9}
\newcommand{\resStepsBExceedShare}{10.3}
\newcommand{\resStepsBExceedText}{35 of 341}
\newcommand{\resStepsBMedian}{9}

\newcommand{\resTabDeviceTime}{{\footnotesize\setlength{\tabcolsep}{3pt}\begin{tabular}{@{}llrrrr@{}}\toprule Benchmark & Configuration & Cells & Attempts & Busy h & Instance h \\ \midrule Cloud-90 & Full system & 180 & 248 & 27.9 & 48.6 \\ Cloud-90 & Single-agent baseline & 180 & 194 & 22.0 & 37.8 \\ Cloud-90 & w/o decomposition (single agent + mechanisms) & 180 & 188 & 21.4 & 32.6 \\ Cloud-90 & - Loop detector & 180 & 198 & 24.2 & 26.0 \\ Cloud-90 & - Text comparison & 180 & 212 & 25.4 & 29.6 \\ Cloud-90 & - Goal verification & 180 & 204 & 23.3 & 25.6 \\ Cloud-90 & - Sequential execution & 180 & 195 & 24.0 & 26.3 \\ Cloud-90 & - Hybrid perception & 180 & 199 & 28.3 & 31.9 \\ Cloud-90 & - Scratchpad memory & 180 & 192 & 25.7 & 28.0 \\ Cloud-90 & All Frontier & 180 & 188 & 22.1 & 25.7 \\ Cloud-90 & Degrade all non-Cortex & 180 & 210 & 23.1 & 26.2 \\ Cloud-90 & Degrade Cortex & 180 & 205 & 69.3 & 76.9 \\ MiniWoB-92 & Full system & 184 & 246 & 18.6 & 24.3 \\ MiniWoB-92 & Single-agent baseline & 184 & 184 & 3.3 & 3.4 \\ MiniWoB-92 & w/o decomposition (single agent + mechanisms) & 184 & 184 & 4.6 & 4.8 \\ \bottomrule\end{tabular}}}
\newcommand{\resTabDeviceTimeNote}{Busy: each cell's own device time, infrastructure retries included. Instance: cloud emulator create-to-delete lifetime (setup and idle included), split across configurations by busy share.}
\newcommand{\nPhys}{16}
\newcommand{\nEmu}{10}

\newcommand{\resAwDecompSig}{S}

\newcommand{\resBudgetSeedLabelCortexBudget}{mean over 2 runs}

\newcommand{\resBudgetSeedLabelNonCortexBudget}{mean over 2 runs}

\newcommand{\resMiniNonIntRewardText}{8 (full system), 12 (w/o decomposition) and 8 (single-agent baseline) of 184 cells per configuration}
\newcommand{\resMiniDestroyedText}{1 cell (w/o decomposition)}
\newcommand{\resMiniInfraLostText}{No cell was lost to infrastructure: every configuration has 2 complete seeds of 92 tasks}
\newcommand{\resLoadControlText}{a 90-task time-matched control at normal host load found no time effect ($\Delta$ = $-5.6$ points, 95\,\% CI [$-12.2$, ${+}1.1$]; exact McNemar $p = 0.18$)}
\newcommand{\resOverloadRerunText}{the 250 cells that had run under host overload were re-run at normal load; 21 changed outcome (14 better, 7 worse; $\Delta$ = ${+}2.8$ points, 95\,\% CI [$-0.8$, ${+}6.4$]; exact McNemar $p = 0.19$)}
\newcommand{\resOssFallbackShare}{1.0\,\%}

\newcommand{\resServedOutsideText}{no call was served outside the allowed set}

\newcommand{\resJudgeModels}{\texttt{openai/gpt-6-astra} and \texttt{anthropic/claude-fable-5.1}}

\newcommand{\resKappaCI}{0.59 (95\,\% CI 0.48--0.70, $n=105$)}
\newcommand{\resDisagreements}{36 of 105}
\newcommand{\resFMReasoningDrift}{10}
\newcommand{\resFMContextPollution}{0}
\newcommand{\resFMStateStaleness}{2}
\newcommand{\resFMTextInputFailure}{9}
\newcommand{\resFMExecutionLatency}{0}
\newcommand{\resFMErrorPropagation}{2}
\newcommand{\resFMRepetitiveLoop}{20}
\newcommand{\resFMPerceptionGrounding}{31}
\newcommand{\resFMTaskMisunderstanding}{7}
\newcommand{\resFMPrematureCompletion}{5}
\newcommand{\resFMEnvironmentOrEvaluator}{6}
\newcommand{\resFMOther}{13}
\newcommand{\resFMCatContextManagement}{12, 11\,\%}
\newcommand{\resFMCatExecutionReliability}{9, 9\,\%}
\newcommand{\resFMCatErrorRecovery}{22, 21\,\%}
\newcommand{\resFMCatOutsideTaxonomy}{62, 59\,\%}

%% file: sections/abstract.tex
\ifwmac
\else
We present \sysfull{}, a multi-agent system for Android UI control, and evaluate it on the 116 tasks of AndroidWorld, on \cloudfirst{}, real Pixel~7 phones from \farm{} and the stock emulator. On fresh, held-out task instances (three pre-registered seeds never used during development), the final system reaches a pass@3 of \resBAWPassAtKtext{}; pass\^{}3 is \resBAWPassHatKtext{} and the mean success rate \resBAWSRtext. \sys{} decomposes each task across five LLM agents (planning, subgoal tracking, state capture, decision, execution) and deterministic nodes, with mechanisms that read typed text back from the device, verify the goal on device evidence and detect repeated failures. A study of the initial system measures what each part contributes: removing only the decomposition changes success by \resDeltaCIDecompOnly{} points, the mechanisms together add \resEffMechanismsCI{} points over a single agent, and the decision agent is the one role that needs a frontier model. We then improve reliability with general changes, a few of them triggered by the goal's wording. On MobileMiniWoB++, a second benchmark never used during development, the final system improves on the initial one by \resBMiniDeltaPts{} points (McNemar $p = \resBMiniDeltaP$; 95\% CI \resBMiniDeltaCIrange). \sys{} is open source: \repo{}
\fi

%% file: sections/intro.tex
\section{Introduction}

Mobile devices are the dominant computing platform, yet autonomous agents remain unreliable at mobile UI control. On the AndroidWorld benchmark, several agentic systems now report success rates above the 80\% human baseline, and five are listed at 100\% on the public leaderboard as of 2026-11-09~\cite{androidworldleaderboard}; the open questions are how reliably, at what cost, and how far beyond this one benchmark such rates hold. Unreliable agents limit applications from automated accessibility testing to hands-free device operation for users with motor impairments.

This paper presents \textbf{\sysfull{}}, a multi-agent system evaluated on the 116 AndroidWorld tasks across 20 applications, on \cloudfirst{}, real Pixel~7 phones from \farm{} and the stock emulator (for the tasks the cloud emulators cannot support: SMS, camera, accessibility hints). On three pre-registered seeds never used during development, the final system reaches a pass@3 of \resBAWPassAtKtext{}, a pass\^{}3 of \resBAWPassHatKtext{} and a mean success of \resBAWSRtext{} (Section~\ref{sec:final}).

We get there in two steps. First, we study the \emph{initial system} (Sections~\ref{sec:failure-modes}--\ref{sec:study}): six failure modes of single-agent architectures, in context management, execution reliability and error recovery, motivate three mechanisms, \emph{cognitive separation} across five LLM agents with focused contexts, \emph{verified execution} that reads text input back from the device, and \emph{meta-reasoning} over the history of agent thoughts; ablations show that the decomposition (\resEffDecompositionCI{} points) and the mechanisms together (\resEffMechanismsCI{}) carry the gain over a single agent, and that only the decision agent needs a frontier model. Second, we improve reliability with general changes developed on other seeds of the same tasks (Section~\ref{sec:reliability}) and evaluate the resulting \emph{final system} on held-out seeds (Section~\ref{sec:final}).

\ifwmac
We use AndroidWorld as a controlled setting to measure three things a task-level score does not show: (1) what decomposing a single agent into cooperating agents adds when the backbone model, tools and harness are held fixed; (2) where model capability has to sit, by downgrading one agent at a time versus all non-central agents together; and (3) what coordination costs, in tokens, latency and LLM calls per agent. A second benchmark of short single-screen tasks (MobileMiniWoB++) shows the other end of the horizon range. Every number comes from repeated runs with confidence intervals.

\fi
\ifaiwild

\fi
We make four contributions:

\begin{enumerate}
    \item \textbf{Failure mode analysis.} We systematically categorize why single-agent mobile automation fails, providing a diagnostic framework for future system design.

    \item \textbf{Multi-agent architecture.} We introduce a decomposition into five LLM agents (Planner, Orchestrator, Contextor, Cortex, Executor) and three deterministic graph nodes (ExecutorToolNode, Summarizer, Convergence), with mechanisms targeting each identified failure mode.

    \item \textbf{A study of the initial system.} Leave-one-out ablations, a comparison without decomposition and a routing study on the Cloud-90 tasks, with two runs per configuration and 95\% CIs, and a second benchmark not used during development (MobileMiniWoB++).

    \item \textbf{Reliability changes and a held-out final evaluation.} General changes, disclosed with their triggers and known limitations, and a pre-registered evaluation on fresh seeds scored by the official evaluator (pass@3, pass\^{}3, mean, with CIs), and a check on MobileMiniWoB++, where the final system improves on the initial one by \resBMiniDeltaPts{} points (McNemar $p = \resBMiniDeltaP$; 95\% CI \resBMiniDeltaCIrange).
\end{enumerate}

We also report the cost of the added architecture: for the initial system, wall-clock time per task on the Cloud-90 tasks is \resWallHeadVsSingle. We release \sys{} as open-source software to support further research in mobile automation.

\section{Related Work}

\textbf{The AndroidWorld Benchmark.} AndroidWorld~\cite{rawles2024androidworld} provides the standard evaluation environment for mobile UI agents, comprising 116 tasks across 20 applications with programmatic success verification. As of 2026-11-09, the public AndroidWorld leaderboard~\cite{androidworldleaderboard} lists several agentic systems above the 80\% human baseline, five of them at 100\% (AGI-0's own report of 2025-10-08 gives 97.4\%~\cite{agio2025}), all self-reported from a single trial. End-to-end models reach up to 80.2\% (MobileRL-9B~\cite{xu2025mobilerl}, listed as AutoGLM-Mobile on the leaderboard). These figures are reported by their authors or the leaderboard, not re-run by us. See Appendix~\ref{sec:extended-related-work} for additional related work on web/desktop agents, reinforcement learning, and benchmarks.

\textbf{Multi-Agent Architectures.} To address error compounding, modular systems decompose mobile interaction into specialized roles, which tend to outperform more general systems~\cite{goldfeder2026ai}. Mobile-Agent~\cite{junyangwang2024mobileagent,junyangwang2024mobilev2} separates planning, perception, and execution. Mobile-Agent-v3~\cite{ye2025mobileagentv3} builds a multi-agent framework on the GUI-Owl model (73.3\%), and Surfer~2~\cite{andreux2025surfer2} combines decoupled planning and execution with self-verification and adaptive recovery (87.1\%). The closest published mechanism to our loop detector is VLAA-GUI's loop breaker~\cite{han2026vlaagui}, which switches interaction mode after repeated failures and forces a strategy change when screen states recur; it is evaluated on desktop benchmarks only. Surveys~\cite{wu2024survey,wang2024survey} note a trend toward hybrid systems combining planning, memory, and verification. \sys{} extends this direction with five LLM agents and three targeted mechanisms for the failure modes we identify.

%% file: sections/problem.tex
\section{Problem Setting}

\subsection{The AndroidWorld Benchmark}

AndroidWorld~\cite{rawles2024androidworld} comprises \textbf{116 programmatic tasks} spanning \textbf{20 real-world Android applications} (calendar, maps, contacts, media, expenses, tasks). Each task is generated from a template with random parameters (e.g., dates, contact names). This prevents memorizing specific task instances. We also report a second benchmark not used during the development of the system (Section~\ref{sec:second-benchmark}).

Success is checked programmatically against device state. Every AndroidWorld result in this paper is scored by the unmodified official evaluator (\texttt{google-research/android\_world}, commit \texttt{55ddaf5}): a task counts as solved according to AndroidWorld's official success check; we never grade a task ourselves, and any deviation from the official setup is listed in Appendix~\ref{app:protocol}.

\subsection{Failure Mode Analysis}
\label{sec:failure-modes}

To understand why prior systems fail, we analyze our first attempt, a single-agent baseline where one LLM handles planning, state assessment, action selection, and execution. The six failure modes below, in three categories, come from our original analysis of this baseline and shaped the design. All counts are re-measured on the re-run baseline (same model and harness as the full system): it reaches \resSingleAgentSRtext{} on the Cloud-90 tasks, and every failed task is labelled with one failure mode by two LLM judges from different model families (\resJudgeModels), given the task goal, the agent trajectory and the evaluator verdict. We report their agreement (Cohen's $\kappa$ = \resKappaCI); an author reviews every disagreement and sets the final label (\resDisagreements{} disagreements).

\textbf{Context management} (\resFMCatContextManagement): \emph{reasoning drift}, where the agent loses track of completed steps in long tasks (\resFMReasoningDrift); \emph{context pollution}, where planning traces mixed with execution logs in one conversation degrade reasoning (\resFMContextPollution); and \emph{state staleness}, where decisions on an outdated UI hierarchy target the wrong element (\resFMStateStaleness). These motivate the multi-agent decomposition, in which each agent keeps a focused context (Section~\ref{sec:multi-agent}).

\textbf{Execution reliability} (\resFMCatExecutionReliability): \emph{text input failures}, where keyboard state or character encoding breaks form filling while the agent believes the text was entered (\resFMTextInputFailure); and \emph{execution latency}, where per-action latency of high-level automation frameworks (\resToolLatencyMs) leaves fewer retries before the time limit (\resFMExecutionLatency). These motivate sequential execution with post-validation, which reads text input back from the device (Section~\ref{sec:post-validation}).

\textbf{Error recovery} (\resFMCatErrorRecovery): \emph{error propagation}, where API failures or unexpected UI states end the task without a recovery attempt (\resFMErrorPropagation); and \emph{repetitive loops}, where the agent repeats the same failing action (\resFMRepetitiveLoop). These motivate meta-reasoning over the history of agent thoughts (Section~\ref{sec:meta-cognitive}).

\textbf{Outside the taxonomy} (\resFMCatOutsideTaxonomy): failures that fit none of these modes, the largest group in the re-run baseline: perception or grounding errors (\resFMPerceptionGrounding), task misunderstanding (\resFMTaskMisunderstanding), premature completion (\resFMPrematureCompletion), environment or evaluator issues (\resFMEnvironmentOrEvaluator), and other (\resFMOther). Two modes that shaped the design, context pollution and execution latency, do not appear in the re-run baseline's labels.

%% file: sections/architecture.tex
\section{System Architecture}
\label{sec:architecture}

\begin{figure*}[t]

\centering
% v1 figure (high_level_architecture2.png) replaced by a TikZ redraw that matches the code.
\resizebox{\ifaiwild0.93\fi\textwidth}{!}{\input{fig_architecture.tex}}
\vspace{0cm}

\caption{\textbf{\sys{}'s execution graph (LangGraph nodes and edges of the released code), with the Cortex as the central decision agent.} (1) Plan: the Planner decomposes the goal into ordered subgoals and the Orchestrator sets their statuses. (2) Act: the Contextor captures device state, the Cortex decides the next actions and/or marks subgoals for review, the Executor turns decisions into tool calls, and the ExecutorToolNode runs them in order on the device, aborting after a failure. The Summarizer trims the message history. Convergence waits for both branches, then routes from the subgoal statuses: continue, replan if a subgoal failed, or end when all are completed. Filled boxes call an LLM; dashed boxes are deterministic. Dashed arrows are taken only in some cycles (replan, subgoal review, no tool call).}
\label{fig:architecture}
\end{figure*}

Figure~\ref{fig:architecture} shows the execution graph. Sections~\ref{sec:multi-agent}--\ref{sec:meta-cognitive} give one mechanism per failure category of Section~\ref{sec:failure-modes} (decomposition, verified execution, meta-reasoning); supporting mechanisms are in Appendix~\ref{app:system}.

\subsection{Multi-Agent Decomposition}
\label{sec:multi-agent}

\sys{} replaces monolithic agent architectures with a modular system built on LangGraph~\cite{langgraph2024}, comprising eight graph nodes and three utility agents. Each component works within a focused context instead of one model holding planning context, perceptual state and execution history at once.

\textbf{Terminology.} We call \emph{agent} only a component that calls an LLM. The graph has eight nodes: five agents (Planner, Orchestrator, Contextor, Cortex, Executor) and three deterministic nodes (ExecutorToolNode, Summarizer, Convergence). Three utility agents (Hopper, Outputter, Video Analyzer; Appendix~\ref{app:system}) are invoked on demand outside the graph. Names match the released code (\codepath{}). The error-recovery mechanism is called \emph{meta-reasoning} throughout.

\subsubsection{Agent Specifications}

\textbf{Planner.} Decomposes the goal into an ordered sequence of subgoals with unique identifiers; it runs at task start and again when a subgoal fails.

\ifbmoved\else\input{sections/arch_orchestrator}\fi

\ifbmoved\else\input{sections/arch_contextor}\fi

\textbf{Cortex.} The central decision agent: from the current subgoal and the device state (UI hierarchy and screenshot together) it produces structured decisions for the Executor, marks subgoals complete for the Orchestrator to review, or both in the same cycle.

\textbf{Executor.} An LLM with the device tools bound that turns the Cortex's structured decisions into tool calls (targets, parameters, and a reasoning annotation).

\ifbmoved\else\input{sections/arch_toolnode}\fi

\textbf{Summarizer.} A deterministic node (despite its name, it calls no LLM) that manages message history to prevent context overflow. When the message count exceeds a threshold (25 messages), it removes older messages while preserving recent context.

\ifbmoved The Orchestrator, the Contextor, the ExecutorToolNode and the execution flow are described in Appendix~\ref{app:components}.\fi

\ifbmoved\else\input{sections/arch_flow}\fi

\subsection{Verified Execution}
\label{sec:post-validation}

Text input represents a disproportionate source of failures in mobile automation. Keyboard state management, focus detection, character encoding, and timing issues combine to produce unreliable text entry. Systems that execute actions without verification risk compounding errors: if action 1 fails silently, actions 2--5 execute against incorrect state.

\sys{} executes the Executor's tool calls strictly sequentially: the ExecutorToolNode aborts the remaining calls of a turn as soon as one fails. Text input goes through a dedicated tool that focuses the field (by resource ID, falling back to coordinates, then text), types, reads the field back when it is addressed by resource ID, and reports both the intended and the actual text to the Cortex (steps in Appendix~\ref{app:system}).

Without the deterministic comparison below, the comparison between intended and actual text is left to the Cortex: the tool returns both strings but does not compare them. The text input logic is separated from LLM reasoning: the Cortex specifies only the target element (with multiple selector types for fallback) and the intended text, and code handles focus, cursor placement, typing, and read-back.

\textbf{Deterministic comparison.} After each successful text input, the tool reads the field back and appends a verdict (MATCH, CONTAINS, CASE\_ONLY, MISMATCH, or UNVERIFIABLE) to the result the Cortex reads, comparing normalized strings (details in Appendix~\ref{app:system}). It is on in the full configuration (flag \texttt{use\_text\_match\_check}); Table~\ref{tab:ablation} reports the system without it. We log every validation event (intended vs.\ read-back text, mismatch, recovery) and report the mismatch rate and how often a mismatch is followed by a successful recovery: \resTextCheck.

\textbf{Goal verification.} Before the episode ends, a goal verifier uses the Cortex model to check every requirement of the goal against fresh device evidence, counting uncommitted effects and partially covered requirements as unmet; an unmet requirement triggers a replan, at most twice per task (details in Appendix~\ref{app:system}; flag \texttt{use\_goal\_verification}; ablated in Table~\ref{tab:ablation}).

\subsection{Meta-Reasoning}
\label{sec:meta-cognitive}

Agents often attempt the same unsuccessful strategy repeatedly without adaptation. For example, an agent may scroll to find a non-existent element multiple times, exhausting available actions without progress. Adaptation is a critical part of generalizable solutions and must be part of the pipeline \cite{goldfeder2026grounded}.

\sys{} addresses this through meta-reasoning in the Cortex. Meta-reasoning itself is implemented at the prompt level; the programmatic loop detector below complements it. Before each decision, the Cortex receives the accumulated history of agent thoughts (its own past decisions, Executor tool outcomes, Orchestrator and Contextor messages). Its system prompt tells it to review this history before acting, to change strategy after repeated failures rather than retry blindly, never to repeat a failed action without understanding why, and to mark subgoals complete only on observed evidence. The Orchestrator's prompt asks it to request replanning when repeated failures make the current plan unworkable.

No numeric threshold exists besides the 400-step limit: recognizing a loop is left to the Cortex. The design draws on CoALA~\cite{sumers2023coala}.

\textbf{Programmatic loop detector.} A deterministic detector tracks (screen state, executed action) transitions and triggers on repeated state--action pairs, on actions that leave the screen unchanged, and on short cycles between few screens (thresholds in Appendix~\ref{app:system}). A trigger adds a loop warning with its evidence to the Cortex input; a second trigger without reaching a new screen marks the current subgoal failed, which forces replanning. It is on in the full configuration (flag \texttt{use\_loop\_detector}) and ablated in Table~\ref{tab:ablation}; triggers: \resLoopTriggers.

\subsection{Evaluation Infrastructure}
\label{sec:eval-infra}

Each task runs in exactly one of three environments: cloud emulators (Android~14) through a non-root port of the task setup; Pixel~7 phones (Android~17) from \farm{}, for tasks that need a settable clock or real Bluetooth; and the stock AndroidWorld emulator, for the tasks the cloud emulators cannot support (SMS, camera, accessibility hints). Each set of runs gives its own counts per environment (Sections~\ref{sec:study} and~\ref{sec:final}; details in Appendix~\ref{app:protocol}). For each task, the official AndroidWorld code initializes the task, the agent receives the goal and controls the device over ADB, and the official evaluator checks success on the device. The initial system is the released code (\repo{}) at commit \agentsha{}, and the final system is frozen \finalref{}; infrastructure failures are handled as in Section~\ref{sec:protocol}.

%% file: fig_architecture.tex
% Figure 1: Minitap execution graph, redrawn 2026-11-09 from the code
% (repos/mobile-use, tag v3.4.0, minitap/mobile_use/graph/graph.py: get_graph()).
% Nodes and edges are exactly the LangGraph nodes and edges:
%   START -> planner -> orchestrator -> convergence
%   convergence --continue--> contextor ; --replan--> planner ; --end--> END
%   contextor -> cortex
%   cortex --review_subgoals--> orchestrator ; --execute_decisions--> executor
%   executor --invoke_tools--> executor_tools ; --skip--> summarizer
%   executor_tools -> summarizer -> convergence
% Filled boxes call an LLM; white dashed boxes are deterministic (no LLM).
\begin{tikzpicture}[
    font=\sffamily\footnotesize,
    >={Stealth[length=2.2mm]},
    every node/.style={align=center},
    agent/.style={draw=violet!70!black, fill=violet!12, rounded corners=2pt,
                  minimum width=2.9cm, minimum height=0.95cm, inner sep=2pt},
    central/.style={agent, line width=1.4pt, fill=violet!25},
    planagent/.style={agent, draw=red!60!black, fill=red!10},
    det/.style={draw=black!70, dashed, fill=white, rounded corners=2pt,
                minimum width=2.9cm, minimum height=0.95cm, inner sep=2pt},
    term/.style={draw=black!80, fill=black!80, text=white, rounded corners=8pt,
                 minimum width=1.5cm, minimum height=0.7cm},
    gate/.style={draw=black!70, dashed, fill=white, diamond, aspect=2.2,
                 inner sep=1pt, minimum width=2.9cm},
    lbl/.style={font=\sffamily\scriptsize, fill=white, inner sep=1pt},
    flow/.style={->, thick},
]
% ---------------- nodes ----------------
\node[term] (start) at (0,0) {START\\[-1pt]\scriptsize user goal};
\node[planagent] (planner) at (3.3,0) {\textbf{Planner}\\\scriptsize goal $\rightarrow$ ordered subgoals};
\node[planagent] (orch) at (7.4,0) {\textbf{Orchestrator}\\\scriptsize sets subgoal statuses};
\node[gate] (conv) at (11.9,0) {\textbf{Convergence}\\\scriptsize joins branches, routes};
\node[term] (end) at (16.0,0) {END};

\node[agent] (ctx) at (11.9,-1.9) {\textbf{Contextor}$^{*}$\\\scriptsize screenshot, UI tree, app, date};
\node[central] (cortex) at (7.4,-1.9) {\textbf{Cortex}\\\scriptsize decides next actions};
\node[det] (summ) at (16.0,-1.9) {\textbf{Summarizer}\\\scriptsize trims message history};

\node[agent] (exec) at (7.4,-3.8) {\textbf{Executor}\\\scriptsize decisions $\rightarrow$ tool calls};
\node[det, minimum width=3.6cm] (tools) at (11.9,-3.8) {\textbf{ExecutorToolNode}\\\scriptsize runs calls in order, aborts on failure};

\node[draw=teal!60!black, fill=teal!10, rounded corners=2pt, minimum width=6.6cm, minimum height=0.7cm]
      (dev) at (11.9,-5.75) {\textbf{Device controller}\ \scriptsize (UIAutomator2 / ADB, idb, WebDriverAgent)};

% ---------------- phase backgrounds ----------------
\begin{scope}[on background layer]
  \node[draw=red!50!black, fill=red!3, rounded corners=6pt, inner sep=6pt,
        fit=(planner)(orch), label={[font=\sffamily\small\bfseries, anchor=south west]north west:Plan}] {};
  \node[draw=violet!50!black, fill=violet!3, rounded corners=6pt, inner sep=6pt,
        fit=(cortex)(ctx)(summ)(exec)(tools), label={[font=\sffamily\small\bfseries, anchor=south west]north west:Act}] {};
\end{scope}

% ---------------- edges ----------------
\draw[flow] (start) -- (planner);
\draw[flow] (planner) -- (orch);
\draw[flow] (orch) -- (conv);
\draw[flow] (conv) -- node[lbl, above] {end} (end);
\draw[flow] (conv) -- node[lbl, right] {continue} (ctx);
\draw[flow, dashed] (conv.north) |- ++(0,0.75) -| node[lbl, pos=0.25, above] {replan (a subgoal failed)} (planner.north);
\draw[flow] (ctx) -- (cortex);
\draw[flow, dashed] (cortex) -- node[lbl, right, align=center] {review\\subgoals} (orch);
\draw[flow] (cortex) -- node[lbl, left, align=center] {execute\\decisions} (exec);
\draw[flow] (exec) -- node[lbl, above] {tool calls} (tools);
\draw[flow] (tools.east) -| (summ.south);
\draw[flow, dashed] (exec.south) -- ++(0,-0.75) -| node[lbl, pos=0.2, below] {no tool call} (summ.south);
\draw[flow] (summ) -- (conv);
\draw[<->, thick, teal!60!black] (tools) -- (dev);

% ---------------- legend ----------------
\node[agent, minimum width=0.5cm, minimum height=0.32cm] (lg1) at (0.0,-2.9) {};
\node[anchor=west, font=\sffamily\scriptsize] at (lg1.east) {LLM agent};
\node[central, minimum width=0.5cm, minimum height=0.32cm] (lg2) at (0.0,-3.4) {};
\node[anchor=west, font=\sffamily\scriptsize] at (lg2.east) {central decision agent};
\node[det, minimum width=0.5cm, minimum height=0.32cm] (lg3) at (0.0,-3.9) {};
\node[anchor=west, font=\sffamily\scriptsize] at (lg3.east) {deterministic (no LLM)};
\node[anchor=west, font=\sffamily\scriptsize, align=left, text width=3.6cm] at (-0.3,-4.65)
     {$^{*}$capture is deterministic; its LLM runs only for app-lock checks};
\end{tikzpicture}

%% file: sections/arch_orchestrator.tex
% moved to Appendix C in the AIWILD version-B build (\ifbmoved); input in both places so the text stays word for word
\textbf{Orchestrator.} Maintains the subgoal statuses (pending, in progress, completed, failed). It examines subgoals marked for completion by the Cortex, decides whether to advance to the next subgoal or mark the current one failed to request replanning, and updates subgoal statuses accordingly. These statuses drive routing: the Convergence gate reads them to decide whether execution continues, terminates, or returns to the Planner.

%% file: sections/arch_contextor.tex
% moved to Appendix C in the AIWILD version-B build (\ifbmoved); input in both places so the text stays word for word
\textbf{Contextor.} Captures the device state before each cycle (screenshot, UI hierarchy, focused application, date and time). Capture is deterministic; the Contextor calls its LLM only when app-lock mode is enabled and the foreground application differs from the locked one, to decide whether to relaunch it.

%% file: sections/arch_toolnode.tex
% moved to Appendix C in the AIWILD version-B build (\ifbmoved); input in both places so the text stays word for word
\textbf{ExecutorToolNode.} A custom LangGraph ToolNode that executes tool calls \emph{sequentially} rather than in parallel. If a call fails, the remaining calls are aborted, so no action runs against stale device state. The node interfaces with a unified controller layer that abstracts platform-specific automation details (Appendix~\ref{app:system}).

%% file: sections/arch_flow.tex
% moved to Appendix C in the AIWILD version-B build (\ifbmoved); input in both places so the text stays word for word
\subsubsection{Execution Flow}

After the Planner and Orchestrator initialize the plan, each cycle runs Contextor $\rightarrow$ Cortex, then branches on the Cortex output (Figure~\ref{fig:architecture}): subgoals marked complete go to the Orchestrator for review, and decisions go to the Executor and ExecutorToolNode, then the Summarizer. Both branches may run in the same cycle. Convergence waits for them, then a deterministic gate reads the subgoal statuses: a failed subgoal returns control to the Planner, all subgoals completed (or none running) hands over to the goal verifier (Section~\ref{sec:post-validation}), and otherwise a new cycle starts.

%% file: sections/experiments.tex
% Table bodies and captions are macros so each build target can lay them out
% (AIWILD/NeurIPS is single-column: side captions and side-by-side tables).
\newcommand{\TabMainAgentRows}{\textbf{\sys{} (Ours)} & Agent & R & \resBAWPassAtK \\
FluizAI\dag & Agent & C & 100.0 \\
Kirk Engine\dag & Agent & C & 100.0 \\
MobileRun & Agent & C & 100.0 \\
AGI-0\dag & Agent & C & 100.0$^a$ \\
agent-qa (Vostride) & Agent & C & 100.0 \\
Artemis & Agent & C & 99.1 \\
MobileUseAgent$^d$ & Agent & C & 97.4 \\
Finalrun & Agent & C & 97.4 \\
askui AndroidVisionAgent & Agent & C & 94.8 \\
AutoDevice & Agent & C & 94.8 \\
Midscene.js & Agent & C & 93.1 \\
}
\newcommand{\TabMainModelRows}{AutoGLM-Mobile (9B)\dag$^e$ & Model & C & 80.2 \\
MAI-UI (235B)\dag & Model & C & 76.7 \\
MAI-UI (32B)\dag & Model & C & 73.3 \\
MAI-UI (8B) & Model & C & 70.7 \\
Gemini 2.5 Computer Use\dag & Model & C & 69.7 \\
GUI-Owl-7B (7B)\dag & Model & C & 66.4 \\
UI-Venus (72B) & Model & C & 65.9 \\
Seed1.5-VL (20B) & Model & C & 62.1 \\
}
\newcommand{\TabMain}{\small
\setlength{\tabcolsep}{3.5pt}
\begin{tabular}{@{}llcr@{}}
\toprule
\textbf{Method} & \textbf{Type} & \textbf{Src.} & \textbf{SR (\%)} \\
\midrule
\multicolumn{4}{l}{\textit{Agentic Frameworks}} \\
\midrule
\TabMainAgentRows\midrule
\multicolumn{4}{l}{\textit{End-to-End Models}} \\
\midrule
\TabMainModelRows\midrule
Human & --- & C & 80.0 \\
\bottomrule
\end{tabular}}
% AIWILD (single column): two panels side by side
\newcommand{\TabMainA}{\small
\setlength{\tabcolsep}{3.5pt}
\begin{tabular}{@{}llcr@{}}
\toprule
\textbf{Method} & \textbf{Type} & \textbf{Src.} & \textbf{SR (\%)} \\
\midrule
\multicolumn{4}{l}{\textit{Agentic Frameworks}} \\
\midrule
\TabMainAgentRows\bottomrule
\end{tabular}}
\newcommand{\TabMainB}{\small
\setlength{\tabcolsep}{3.5pt}
\begin{tabular}{@{}llcr@{}}
\toprule
\textbf{Method} & \textbf{Type} & \textbf{Src.} & \textbf{SR (\%)} \\
\midrule
\multicolumn{4}{l}{\textit{End-to-End Models}} \\
\midrule
\TabMainModelRows\midrule
Human & --- & C & 80.0 \\
\bottomrule
\end{tabular}}
\newcommand{\CapMain}{\caption{\textbf{AndroidWorld results (116 tasks).} SR (\%): for other systems, the success rate as reported by each source; for ours, pass@3 over 3 held-out seeds (Section~\ref{sec:final}). \emph{Agent}: multi-component frameworks; \emph{Model}: end-to-end VLM (one model call per step). \dag~closed-source. \emph{Src.}: R = run by us under the protocol of Section~\ref{sec:protocol}; C = cited from the public AndroidWorld leaderboard~\cite{androidworldleaderboard}, snapshot of 2026-11-09 (the system's own page or paper: Appendix~\ref{sec:extended-related-work}). Cited rows are self-reported and not independently verified; every one that states a trial count, including all five at 100\%, is a single trial. Rows shown: all agentic entries $\geq$90\% and all end-to-end models $\geq$60\% in the snapshot. $^a$The 100 appears only in a leaderboard note, which also states that its models were trained on AndroidWorld tasks; AGI's own report of 2025-10-08 gives 97.4~\cite{agio2025}. $^d$Leaderboard only; no report found, and the Seed1.8 model card gives 70.7 for the model~\cite{seed2026seed18}. $^e$Reported as MobileRL-9B~\cite{xu2025mobilerl}.}
\label{tab:main_results}}
\newcommand{\TabAbl}{
\begin{tabular}{@{}lcc@{}}
\toprule
\textbf{Configuration} & \textbf{SR} & \textbf{$\Delta$ (95\% CI)} \\
\midrule
Full system (reference) & \resSRcellFull & --- \\
\midrule
Single-agent baseline$^a$ & \resSRcellSingleAgent & \resDeltacellSingleAgent \\
w/o decomposition$^b$ & \resSRcellDecompOnly & \resDeltacellDecompOnly \\
\midrule
\multicolumn{3}{@{}l}{\textit{Component removed}} \\
Loop detector & \resSRcellNoLoop & \resDeltacellNoLoop \\
Text comparison & \resSRcellNoTextCheck & \resDeltacellNoTextCheck \\
Goal verification & \resSRcellNoGoalVerif & \resDeltacellNoGoalVerif \\
Sequential execution & \resSRcellNoSequential & \resDeltacellNoSequential \\
Hybrid perception & \resSRcellNoHybrid & \resDeltacellNoHybrid \\
Scratchpad memory & \resSRcellNoScratchpad & \resDeltacellNoScratchpad \\
\bottomrule
\end{tabular}}
\newcommand{\CapAbl}{\caption{\textbf{Ablation on the AndroidWorld Cloud-90 subset (90 tasks, 2 seeds, paired with the full system).} SR: mean; $\Delta$ vs.\ the full system, 95\% bootstrap CI. $^a$One LLM (Cortex model), every mechanism off incl.\ vision. $^b$One monolithic agent, every other mechanism on (loop detector inactive).}
\label{tab:ablation}}
\newcommand{\TabRout}{
\begin{tabular}{@{}llcc@{}}
\toprule
\textbf{Configuration} & \textbf{Downgraded} & \textbf{SR} & \textbf{\$/task} \\
\midrule
All Frontier & none & \resSRcellAllFrontier & \resCostAllFrontier \\
Degrade all non-Cortex & all but Cortex & \resSRcellNonCortexBudget & \resCostNonCortexBudget \\
\midrule
Platform Default & all but Cortex (mixed) & \resSRcellFull & \resCostFull \\
Degrade Cortex & Cortex too & \resSRcellCortexBudget & \resCostCortexBudget \\
\bottomrule
\end{tabular}}
\newcommand{\CapRout}{\caption{\textbf{Routing study (AndroidWorld Cloud-90 subset, 90 tasks, one frontier model).} SR: mean $\pm$ 95\% CI over two runs; budget rows, Degrade Cortex: \resBudgetSeedLabelCortexBudget; Degrade all non-Cortex: \resBudgetSeedLabelNonCortexBudget. Cost: LLM cost per task at the stated price date. Downgrades use one budget model, Qwen3-VL-8B. Top block: from All Frontier. Bottom block: Platform Default (mixed budget models outside the Cortex, Appendix~\ref{app:efficiency}) and the same with the Cortex downgraded.}
\label{tab:routing}}

\section{Study of the Initial System}
\label{sec:study}

This section studies the system of Section~\ref{sec:architecture} before the reliability changes of Section~\ref{sec:reliability}, which we call the \emph{initial system}. Its ablations and routing study were not re-run on the final system.

\subsection{Evaluation Protocol}
\label{sec:protocol}

\textbf{Statistical protocol.} Every result in this section comes from fresh runs on one pinned agent commit (the initial system), and the final evaluation of Section~\ref{sec:final} follows the same protocol on the final system's frozen commit; no result from earlier runs is reused. The full system and the single-agent baseline run all 116 tasks; the ablations (where the single agent enters through its Cloud-90 tasks), the variant without decomposition and the routing study run on the Cloud-90 tasks, paired with the full system on the same tasks and seeds. Every configuration is run twice (seeds 30 and 31), and every task runs under a 30-minute wall-clock cap and a 400-step limit; a task that hits the cap counts as a failure. Runs are independent: each uses fresh task parameters (a different AndroidWorld \texttt{task\_random\_seed}), a fresh device state, and the same pinned agent commit. Success is decided only by the official AndroidWorld evaluator. Success over the 116 tasks is pooled: each task runs in exactly one environment (90 on cloud emulators, \nPhys{} on phones, \nEmu{} on the stock emulator), so per-task success over the 116 is well defined, and the per-environment breakdown is always reported next to it. Single-trial pass@1 figures and naive aggregation can mislead on agent benchmarks~\cite{doro2026precipice}, so we report the mean success rate with a 95\% confidence interval from a nonparametric bootstrap over tasks and runs, pass@$k$ (a task counts as solved if any of $k$ runs solves it) and pass\^{}$k$ (solved in all $k$ runs), and a per-task flip table; timeouts, infrastructure failures and logging are detailed in Appendix~\ref{app:protocol}.

\subsection{The Initial System on AndroidWorld}
\label{sec:main-results}

On all 116 tasks, the initial system reaches \resHeadlineSRtext{} pooled over the 116 tasks (Cloud-90: \resLimHeadlineSRtext; phones: \resPhysHeadlineSRtext; stock emulator: \resEmuHeadlineSRbrief). Run-to-run reliability: \resHeadlineBrittleText.

\subsection{Ablation Studies}
\label{sec:ablations}

Table~\ref{tab:ablation} quantifies the contribution of each architectural component. Each row removes one component from the full system (the full configuration of Section~\ref{sec:main-results}, Platform Default routing). $\Delta$ is the difference of mean success rates, with a 95\% bootstrap CI over tasks and runs.

\begin{table}[ht!]
\ifaiwild
\begin{minipage}[t]{0.47\linewidth}\CapAbl\centering\fitcol{\TabAbl}\end{minipage}\hfill
\begin{minipage}[t]{0.49\linewidth}\CapRout\centering\fitcol{\TabRout}\end{minipage}
\else
\CapAbl
\centering
\fitcol{\TabAbl}
\fi
\end{table}

Two changes matter: replacing the system by one LLM with every mechanism off (\resDeltaCISingleAgent{} points) and removing only the decomposition (\resDeltaCIDecompOnly{} points). Removing any single other component causes no significant change (every 95\% CI includes zero; Table~\ref{tab:ablation}): once the others are present, no single mechanism has a significant effect on its own on these tasks, although the mechanisms together add \resEffMechanismsCI{} points over the single-agent baseline (variant without decomposition vs.\ single-agent baseline). Leave-one-out deltas are not additive: components interact, so the deltas need not sum to the gap between the single-agent baseline and the full system.

\subsection{Model Routing and Efficiency}
\label{sec:efficiency}

A natural question arises when deploying multi-agent systems: must all agents
use frontier models, or can a heterogeneous model allocation reach comparable
performance at lower cost? We answer it with one routing study on the Cloud-90 tasks (paired runs), using a single frontier model (Gemini~3.1 Pro Preview, \texttt{google/gemini-3.1-pro-preview})
throughout and one budget model (Qwen3-VL-8B) for every downgrade, so that each
configuration differs from \emph{All Frontier} or from \emph{Platform Default} only in which roles are downgraded.

Each configuration runs with two seeds; the models of each role and the methodology are in Appendix~\ref{app:efficiency}.

\ifaiwild\else
\begin{table}[ht!]
\CapRout
\centering
\fitcol{\TabRout}
\end{table}
\fi

\paragraph{Results.}
Platform Default (the full configuration of the initial system) and All Frontier do not differ significantly in success (\resRDeltaCIFull{} points), at \$\resCostFull{} vs.\ \$\resCostAllFrontier{} per task. Downgrading the Cortex alone collapses success to \resSRtextCortexBudget{}, \resPDDeltaCICortexBudget{} points relative to Platform Default. Downgrading all non-Cortex agents at once changes it by \resRDeltaCINonCortexBudget{} points relative to All Frontier, which stays within the CI of All Frontier, at \$\resCostNonCortexBudget{} per task: on these tasks the frontier capability must sit in the Cortex, and downgrading the other agents causes no significant drop. Fig.~\ref{fig:pareto_frontier} (Appendix~\ref{app:efficiency}) shows the cost--success trade-off. We state a routing principle only if the CIs support it; in particular, we do not claim that non-Cortex agents tolerate budget models unless the all-non-Cortex downgrade stays within the CI of All Frontier.

\subsection{Second Benchmark}
\label{sec:second-benchmark}

AndroidWorld is close to saturation for current agents~\cite{kong2026mobileworld}, so we also evaluate on MobileMiniWoB++~\cite{rawles2024androidworld}, a second benchmark not used during the development of the system: the Android port of the MiniWoB++ web tasks~\cite{shi2017wob,liu2018miniwobpp} shipped with AndroidWorld (official reward). Its tasks are short interactions with simple web widgets, not long-horizon app use. The headline is the 62-task subset of the AndroidWorld paper (human 100\%, best 2024 agent 67.7\%~\cite{rawles2024androidworld}), with all 92 tasks as a secondary result. \sys{} runs unchanged on the cloud emulators (Android~14), two seeds each for the full system, the single-agent baseline and the variant without decomposition (Table~\ref{tab:mini}; deviations in Appendix~\ref{app:protocol}). \emph{The system transfers:} the full system reaches \resMiniHeadlineSRtext{} on the 62-task subset (\resMiniHeadlinePassText; all 92 tasks: \resMiniAllHeadlineSRtext), comparable to the best 2024 agent (cited, not run under our protocol). \emph{Mechanisms versus decomposition:} the single-agent baseline (one LLM, every mechanism off) reaches \resMiniSRtextSingleAgent{} and the variant without decomposition (one agent, every mechanism on) \resMiniSRtextDecompOnly{}. The mechanisms add \resMiniEffMechanismsCI{} points over the single agent, while the decomposition changes success by \resMiniEffDecompositionCI{} points ($p = \resMiniEffDecompositionP$), no significant gain on these short tasks, at a higher cost per task (\$\resMiniCostFull{} vs.\ \$\resMiniCostDecompOnly{}; Table~\ref{tab:mini}).

%% file: sections/reliability.tex
\section{Improving Reliability}
\label{sec:reliability}

To improve reliability, we then changed the initial system without changing its models, agents or routing. Every change is stated as a general behaviour and motivated by a class of failures seen in development, never by a task; a few are triggered by the goal's wording. Appendix~\ref{app:versionB} lists every change with the failure class behind it.

\textbf{Rules for the agents.} The Planner and the Cortex read a whole list or source before counting, answering or declaring an item absent, decide a category from the item's field, never from its title, and include an item exactly at a goal's bound unless the goal says ``strictly''; the goal is turned once into explicit requirements.

\textbf{Verification.} The goal verifier judges each requirement on device evidence only (screens, deterministic logs of recent screens and executed actions, tool outputs), treats agent statements as claims, rejects at most 3 times, and its final answer is the reported answer.

\textbf{Execution and tools.} Predictable actions the Cortex decided but the Executor's LLM left out are dispatched in order (executor completion). New read-only tools read a whole list (\texttt{read\_full\_list}), inspect matching rows (\texttt{inspect\_rows}) and compute (\texttt{calculate}); and the internal limits rise to \resBLimitSteps{} graph steps and \resBLimitSeconds\,s per task.

\textbf{Rules triggered by the goal's wording} (disclosed): names with a file extension; records copied from a named source file (two of its three checklist lines are borderline-general); date-window questions, read with the app's full-date display (an exception to ``questions are read-only''); tracks or routes through given places (a fixed planner procedure, borderline-general); and two verifier gates for delete goals.

\textbf{Development protocol.} The changes were developed on development seeds of the AndroidWorld tasks (\resBDevSeeds) over \resBNIterations{} iterations; seeds 30 and 31 are those of the initial system's study, so re-checks on them served only as a development gate. The final evaluation uses three fresh, held-out seeds (\resBFinalSeeds), declared in advance and never used before the freeze, and MobileMiniWoB++ was never used during development. \resBFreezeText{}

\textbf{Known limitations.} In the route procedure, a point can land away from its place and still be accepted by the verifier (observed in \resBRouteRiskText); \texttt{inspect\_rows} does not resume past rows it already reported.

\section{Final Evaluation}
\label{sec:final}

We evaluate the final system on the 116 tasks with the three held-out seeds, \resBK{} runs per task, under the protocol of Section~\ref{sec:protocol}: \resBNCloud{} tasks on cloud emulators, \resBNPhys{} on phones and \resBNEmu{} on the stock emulator (three tasks move to the stock emulator because the cloud image cannot pass them; Appendix~\ref{app:versionB}). The ablations and the routing study of Section~\ref{sec:study} use the initial system; we did not re-run them on the final system.

The final system reaches a pass@\resBK{} of \resBAWPassAtKtext{} (\resSportsFinalAmbigText{}; see Appendix~\ref{app:protocol}), a pass\^{}\resBK{} of \resBAWPassHatKtext{} and a mean success of \resBAWSRtext{}. Per environment, pass@\resBK{} is \resBCloudPassAtK\,\% on cloud emulators, \resBPhysPassAtK\,\% on phones and \resBEmuPassAtK\,\% on the stock emulator, and \resBFreshPassAtKtext{} on the 92 tasks with fresh instances (all CIs and pass\^{}\resBK{} in Table~\ref{tab:versionB}). Run-to-run reliability: \resBAWBrittleText{} (Table~\ref{tab:flipB}). Before the reliability changes, the initial system reached a mean of \resHeadlineSRtext{}, with \resHeadlinePassText, on seeds 30 and 31 (Section~\ref{sec:main-results}); the two sets of runs use different seeds and a different $k$, so we report them side by side and compute no difference. \resBFirstAfterText{}

\textbf{Comparison with the leaderboard} (Table~\ref{tab:main_results}). Cited rows use different backbone models, protocols and numbers of runs (most a single trial), so a gap of a few points between rows is not evidence that one architecture is better. This paper complements single-trial results with fresh, repeated runs of a single frozen version, which measure how reliably each task is solved (pass@k, pass\^{}k).

\begin{table}[ht!]
\ifaiwild
\begin{minipage}[t]{0.47\linewidth}\centering\fitcol{\TabMainA}\end{minipage}\hfill
\begin{minipage}[t]{0.51\linewidth}\centering\fitcol{\TabMainB}\end{minipage}
\CapMain
\else
\CapMain
\centering
\fitcol{\TabMain}
\fi
\end{table}

\textbf{Second benchmark.} On MobileMiniWoB++, never used during development, the final system improves on the initial system by \resBMiniDeltaPts{} points on the 92 tasks (seeds 30 and 31, paired; McNemar $p = \resBMiniDeltaP$, 95\% bootstrap CI \resBMiniDeltaCI{}; 62-task subset: \resBMiniSubsetDeltaCI{}); the pre-registered no-regression criterion (CI lower bound above $-5$ points) is met.

%% file: sections/conclusion.tex
\section{Discussion and Conclusion}

\textbf{Limitations.} AndroidWorld evaluates 20 applications with well-defined task structures; performance on arbitrary applications with novel UI patterns may differ. Element targeting falls back to screen coordinates when no identifier is available, which may break under layout, theme, or display-scale changes. The system relies on UI accessibility hierarchies, so applications without accessibility support (mobile games, canvas-based interfaces) cannot be reliably automated. On short single-screen tasks (MobileMiniWoB++), the multi-agent decomposition adds cost without a significant gain (Section~\ref{sec:second-benchmark}). The reliability changes are general-purpose and were evaluated on held-out seeds and on MobileMiniWoB++, which was never used during development; a residual risk that they favour this benchmark cannot be fully excluded. Each task starts fresh, without cross-task learning.

\ifwmac\textbf{Measuring the collective.} \awdecomp{The two benchmarks disagree on what decomposition adds. On the Cloud-90 tasks it adds \resEffDecompositionCI{} points (Table~\ref{tab:ablation}), at a cost per task of \$\resCostFull{} with vs.\ \$\resCostDecompOnly{} without it. On MobileMiniWoB++'s short single-screen tasks it brings no significant gain, at \$\resMiniCostFull{} vs.\ \$\resMiniCostDecompOnly{}, and the gain over a single agent comes from the other mechanisms (Section~\ref{sec:second-benchmark}). What decomposition adds, in success and in cost, is task-dependent.}{Decomposition helped on neither benchmark at significance level: on MobileMiniWoB++'s short single-screen tasks it adds cost but no significant success (Section~\ref{sec:second-benchmark}), and on the Cloud-90 tasks it changes success by \resEffDecompositionCI{} points (Table~\ref{tab:ablation}); the gains over a single agent come from the mechanisms. Whether a decomposition pays off on longer or more open-ended tasks is an open question.} Capability must sit in the decision agent: downgrading the Cortex alone changes success by \resPDDeltaCICortexBudget{} points, while downgrading all other agents at once changes it by \resRDeltaCINonCortexBudget{} relative to All Frontier, at \$\resCostNonCortexBudget{} vs.\ \$\resCostAllFrontier{} per task. Measuring a sub-agent architecture therefore takes paired single-agent and no-decomposition baselines and per-role downgrades, on more than one benchmark.

\fi
\textbf{Conclusion.} This paper presents \sys{}, a multi-agent system that, after general reliability changes, reaches a pass@\resBK{} of \resBAWPassAtKtext{}, a pass\^{}\resBK{} of \resBAWPassHatKtext{} and a mean success of \resBAWSRtext{} on AndroidWorld on held-out seeds, and improves MobileMiniWoB++ by \resBMiniDeltaPts{} points (McNemar $p = \resBMiniDeltaP$; 95\% CI \resBMiniDeltaCIrange). \awdecomp{On the Cloud-90 tasks (Section~\ref{sec:ablations}), the ablations support one design principle, cognitive specialization (decomposition \resEffDecompositionCI{} points), an effect that does not transfer to the short single-screen tasks of MobileMiniWoB++. The other mechanisms help together (\resEffMechanismsCI{} points over the single-agent baseline), but none has a significant effect on its own once the others are present.}{On the Cloud-90 tasks (Section~\ref{sec:ablations}), neither the decomposition (\resEffDecompositionCI{} points) nor any single other mechanism is significant on its own (text comparison \resDeltaCINoTextCheck{}, loop detector \resDeltaCINoLoop{} points); the mechanisms together change success by \resEffMechanismsCI{} points over the single-agent baseline.} More broadly, system design, not only model capability, sets how reliably current models complete long-horizon mobile tasks; even for the final system, \resBAWFlipN{} of the 116 tasks are solved in some but not all runs.

%% file: sections/impact.tex
\section*{Impact Statement}

Mobile automation agents can improve accessibility for users with motor impairments and reduce tedious manual testing. However, agents with UI access can observe sensitive device information. \sys{} operates only on user-initiated tasks and requires explicit device access. We release it as open-source to enable community scrutiny.

%% file: sections/appendix.tex
\section{Extended Experimental Details}

This appendix provides additional experimental details and supplementary results.

\subsection{Prompt Templates}

The Cortex agent receives instructions emphasizing multimodal integration:

\begin{quote}
\textit{``You have two complementary perception modalities: the UI accessibility hierarchy provides precise element identifiers, coordinates, and structural relationships, the screenshot provides visual context including appearance, spatial layout, and visibility state. You must integrate both modalities, using structural information for element targeting and visual information for context and verification.''}
\end{quote}

The data fidelity instruction addresses over-helpful behavior:

\begin{quote}
\textit{``For any data-related task: transcribe, enter, or reproduce content exactly as specified unless explicitly instructed otherwise. Do not correct formatting, add punctuation, or modify content in any way.''}
\end{quote}

\section{Additional System Details}
\label{app:system}

\ifbmoved
\subsection{Further Components and Execution Flow}
\label{app:components}

\input{sections/arch_orchestrator}
\input{sections/arch_contextor}
\input{sections/arch_toolnode}
\input{sections/arch_flow}
\fi

\subsection{Utility Agents}

Three additional agents operate outside the main execution graph:

\begin{itemize}
    \item \textbf{Outputter}: Generates structured output from task results according to user-specified schemas.
    \item \textbf{Hopper}: Extracts specific data from provided content using targeted prompts.
    \item \textbf{Video Analyzer}: Analyzes screen recordings using video-capable models (e.g., Gemini) for tasks requiring temporal understanding.
\end{itemize}

\subsection{Platform Automation}

\sys{} provides unified automation across platforms through a controller abstraction layer:

\begin{itemize}
    \item \textbf{Android}: UIAutomator2~\cite{uiautomator2} for UI automation with ADB shell commands as fallback
    \item \textbf{iOS Simulators}: fb-idb~\cite{idb} for comprehensive simulator control
    \item \textbf{iOS Physical Devices}: WebDriverAgent~\cite{webdriveragent2015} for physical device automation
\end{itemize}

This architecture enables platform-agnostic task definitions while leveraging native automation capabilities for reliability. The original ablation also replaced UIAutomator2 with Maestro, the previous automation backend. The Maestro integration was removed from the code on 2025-12-03, so that configuration cannot be re-run on the pinned commit, and its row is dropped from Table~\ref{tab:ablation}.

\subsection{Known Defect}

The Cortex receives the device information and date but never the focused-application name (an operator-precedence error in the prompt construction); the defect is present in both the initial and the final system.

\subsection{Goal Verifier}

When the subgoal graph would end, a goal verifier runs instead of terminating immediately. It reads the original goal, the executed plan, recent agent thoughts and a fresh capture of the device (UI hierarchy and, with vision on, a screenshot), and uses the Cortex model to decide whether every requirement of the goal is satisfied by visible evidence. Prepared but uncommitted effects (text typed but not submitted, content selected but not copied, a form not confirmed) and partially covered multi-item requirements count as unmet, and for a goal that asks for information it must state the answer as observed. If a requirement is unmet, it adds a failed subgoal describing what is missing and returns control to the Planner, at most twice per task; otherwise the episode ends.

\subsection{Deterministic Text Comparison and Loop Detector}

Both mechanisms are part of the initial system (commit \agentsha{}), and each is controlled by its own configuration flag.

\textbf{Text comparison.} After a successful text input, the field is read back by resource ID (accepting both hierarchy key formats) or, failing that, as the focused text field. Both strings are normalized (NFKC, zero-width characters removed, non-breaking spaces to spaces, whitespace collapsed and trimmed) and compared. Verdicts: MATCH, CONTAINS, CASE\_ONLY, MISMATCH, or UNVERIFIABLE (password field or no field found). Each check is logged (intended and read-back text, verdict, read-back and focus method).

\textbf{Loop detector.} At each Cortex turn the detector closes one transition, previous state $\rightarrow$ executed tool calls $\rightarrow$ new state. A state is a hash of the focused application and of the resource ID, class, text, content description, and checked/selected/focused/enabled flags of every UI element (system UI dropped, clock strings masked, bounds ignored). Triggers (default thresholds): the same state--action pair $\geq$3 times in the last 12 transitions; 3 consecutive UI actions without screen change (note and wait tools excluded); $\leq$2 distinct screens over the last 8 transitions. After a trigger, a 3-transition cooldown applies; after 2 triggers without reaching a never-seen screen, the current subgoal is marked failed and control goes straight to the Planner. Every trigger is logged.

\subsection{Supporting Mechanisms}
\label{sec:supporting}

\textbf{Hybrid perception.} The Cortex receives both the UI hierarchy and a screenshot at every decision, using structural information for element targeting and visual information for appearance and context. \textbf{Scratchpad memory.} A persistent key-value scratchpad (\texttt{save\_note}, \texttt{read\_note}, \texttt{list\_notes}) keeps information across the whole task. Video reasoning is described below.

\subsection{Text Input with Post-Validation}

The text-input tool runs these steps: (1) focus the field by resource ID, falling back to coordinates, then text matching; (2) move the cursor to the end of existing content when bounds are known; (3) type the text through the platform automation layer; (4) if the field is addressed by resource ID, read a fresh UI hierarchy and extract the field's current text; and (5) report both the intended text and the field's content to the Cortex. Without a resource ID there is no read-back, and the tool result asks the Cortex to verify on the next screen.

\subsection{Video-Based Temporal Reasoning}
Static screenshots capture only instantaneous state. Certain tasks require temporal observation: verifying animations complete correctly, observing multi-step UI transitions, or confirming video playback behavior. \sys{} provides \texttt{start\_video\_recording()} and \texttt{stop\_video\_recording(prompt)} operations that capture and analyze temporal sequences using a vision-language model.

\section{Extended Related Work}
\label{sec:extended-related-work}

This appendix provides additional context on related work not essential to the main narrative.

\subsection{Error Recovery and Robustness}

Reliable agents require mechanisms to detect and correct mistakes. InfiGUI-R1~\cite{liu2024infiguir1} trains reasoning with spatial reasoning distillation, then reinforcement learning with sub-goal guidance and constructed error-recovery scenarios. AppAgentX~\cite{jiang2025appagentx} develops memory mechanisms to identify repetitive action sequences. These approaches motivate our meta-reasoning mechanism, in which the decision agent reviews the history of agent thoughts to detect repeated failures and change strategy. BacktrackAgent~\cite{wu2025backtrackagent} detects errors in GUI agents and recovers by undoing actions rather than by changing strategy. DeLoop~\cite{chen2026deloop} defines a loop as a repeated tool call that returns the same observation and reduces such loops by preference training (from 57.3\% to 6.1\% of Qwen3-8B tasks); it studies text agents, not GUI agents, and fixes the problem in training rather than detecting it at run time as our loop detector does.

\subsection{Agentic AI for Web and Desktop Applications}

LLM-based agents now control computers through natural language. The ReAct paradigm~\cite{yao2022react} established the pattern of interleaving reasoning with action; subsequent systems extended this to browsers and operating systems. Humphreys et al.~\cite{humphreys2022control} showed that agents can learn desktop control directly from pixel demonstrations. Vision-language models have since enabled interaction with arbitrary GUIs by grounding UI elements in visual input~\cite{kim2024llmsolve,zhang2024ufo,wu2024oscopilot,wu2024osatlas}. Agent S~\cite{agashe2024agents} combines experience-augmented planning with a specialized agent-computer interface, reaching 20.6\% on OSWorld, an 83.6\% relative improvement over the best baseline.

Benchmarks measure this progress. WebShop~\cite{yao2023webshop}, WebArena~\cite{zhou2023webarena}, and Mind2Web~\cite{deng2023mind2web} evaluate web-based tasks. OmniACT~\cite{kapoor2024omniact} extends evaluation to desktop and web workflows, and UINavBench~\cite{agrawal2025uinavbench} to multi-step tasks on mobile interfaces.

\subsection{Additional Mobile Benchmarks}

\textbf{Sources of Table~\ref{tab:main_results}.} Besides the public leaderboard, each system's own page or paper where one exists: FluizAI~\cite{fluizai2026}, Kirk Engine~\cite{kirkengine2026}, MobileRun~\cite{mobilerun2026}, agent-qa~\cite{vostride2026agentqa}, Artemis~\cite{artemis2026}, Finalrun~\cite{finalrun2026}, Midscene.js~\cite{midscene2026}, MAI-UI~\cite{zhou2025maiui}, GUI-Owl~\cite{ye2025mobileagentv3}, UI-Venus~\cite{gu2025uivenus}, Seed1.5-VL~\cite{guo2025seed15vl}.

MobileWorld~\cite{kong2026mobileworld} argues that AndroidWorld is saturated, with agents above 90\%, and proposes a harder successor (201 tasks, 62.2\% of them spanning several apps) on which the best agentic framework reaches 51.7\%; AndroidLab~\cite{xu2025androidlab} and MobileAgentBench~\cite{wang2024mobileagentbench} are other reproducible Android suites.

Beyond AndroidWorld, several benchmarks evaluate mobile agents. Android in the Wild~\cite{rawles2024androidwild} provides large-scale demonstrations of real user behavior. The A3: Android Agent Arena~\cite{chai2025a3} benchmark introduced essential-state procedural evaluation using multimodal language models for progressive verification across 100 tasks from dynamic online applications. Recent surveys~\cite{liu2025phonesurvey} provide taxonomies of agent frameworks, modeling approaches, and evaluation methodologies in phone automation. Early work on mobile instruction following focused on mapping language to UI actions~\cite{li2020acl}, interactive feedback~\cite{burns2021motif}, and grounded instruction following~\cite{venkatesh2022ugif}. AndroidReality~\cite{liu2026androidreality} perturbs AndroidWorld tasks and finds substantial robustness gaps that clean success rates hide. On MobileWorld~\cite{kong2026mobileworld}, Qwen-UI-Agent~\cite{zhou2026qwenuiagent} reports 82.1\% and GUI-Owl-1.5~\cite{xu2026mobileagentv35} 46.8\% (71.6\% on AndroidWorld). Reliable centralized benchmarking is one of the main drivers of scientific progress \cite{goldfeder2026driving}, and we expect further progress to accelerate as benchmarking efforts improve and coalesce. 

\subsection{Reinforcement Learning Approaches}

Reinforcement learning methods fine-tune agents through online interaction with devices. DigiRL~\cite{bai2024digirl} demonstrated that combining offline and online RL can improve success rates from 17.7\% to 67.2\% on Android-in-the-Wild. Digi-Q~\cite{bai2025digiq} learns Q-value functions using frozen VLM features with offline temporal-difference learning, achieving 21.2\% improvement over prior methods. MobileGUI-RL~\cite{shi2025mobileguirl} advances mobile GUI agents through RL in online environments, while VSC-RL~\cite{wu2025vscrl} reformulates decision-making as variational subgoal-conditioned RL. ARPO~\cite{lu2025arpo} introduces end-to-end policy optimization with experience replay, and ZeroGUI~\cite{yang2025zerogui} enables automated online learning without human annotation costs. GiGPO~\cite{feng2025gipo} proposes group-in-group policy optimization for more efficient LLM agent training.

\subsection{Additional End-to-End and Foundation Models}

Vision-language models can directly map screenshots to actions. UI-TARS~\cite{qin2025uitars} incorporates task decomposition and reflection, while UI-Venus~\cite{gu2025uivenus} uses reinforcement finetuning to reach 65.9\% on AndroidWorld with its 72B variant, and MAI-UI~\cite{zhou2025maiui} reaches 76.7\% (235B) while routing between on-device and cloud models, which cuts cloud calls by over 40\%. Over long horizons, a single wrong action can derail such end-to-end agents.

Systems such as CogAgent~\cite{hong2024cogagent} and AppAgent~\cite{zhang2023appagent} demonstrate that multimodal models can act as smartphone users by interpreting screenshots and issuing UI actions. Ferret-UI 2~\cite{li2025ferretui2} extended this paradigm to universal interface understanding across iPhone, Android, iPad, web, and AppleTV platforms through adaptive scaling and enhanced training data generation. Android in the Zoo~\cite{zhang2024androidzoo} explores chain-of-thought-style reasoning for GUI agents. MobileUse~\cite{li2025mobileuse} introduces hierarchical reflection mechanisms for improved reasoning at multiple abstraction levels. UI-TARS-2~\cite{wang2025uitars2} raises AndroidWorld success from 64.6\% to 73.3\% with reinforcement learning. Among agentic systems not shown in Table~\ref{tab:main_results}, K$^2$-Agent~\cite{wu2026k2agent} uses a hierarchical reasoner and executor (76.1\% in its paper; the leaderboard lists 76.7\%), and AgentProg~\cite{tian2025agentprog} manages long-horizon context with an execution tree (78.0\%, 61.6\% without it). V-Droid~\cite{dai2025vdroid} adds pre-execution verification, achieving 59.5\% success on AndroidWorld with 6.1$\times$ lower per-step latency (4.3\,s/step). SPlanner~\cite{mo2025splanner} models application control flows with finite state machines, reaching 63.8\% when paired with Qwen2.5-VL-72B.

\subsection{Efficiency and Deployment}

Efficiency considerations have led to approaches like LLM-Explorer~\cite{zhao2025llmexplorer}, which achieves 148$\times$ lower cost than prior LLM-based methods, and AutoDroid-V2~\cite{wen2025autodroidv2}, which enables on-device deployment through code generation with smaller language models. Recent work also addresses agent-initiated interaction~\cite{kahlon2025interaction}, recognizing that complex tasks often require proactive user engagement. LaMo~\cite{wang2026lamo} pairs a 3B executor with stronger planners, and Jev-Mobile~\cite{zhang2026jevmobile} decouples planning from execution, giving up about 5 points on AndroidWorld (79\% vs.\ 84\%) for 73.4\% lower API cost on successful trajectories. Model routing is usually studied per query: RouteLLM~\cite{ong2025routellm} learns a strong/weak router from preference data, and FrugalGPT~\cite{chen2024frugalgpt} cascades models by cost. AgentRouter~\cite{paul2026agentrouter} routes each step of an agent trajectory to one of four model tiers (72\% lower cost at 97.3\% of frontier quality, on planning, coding, research and data-analysis tasks). We found no study of per-agent-role model mixes for mobile GUI agents; our routing study (Section~\ref{sec:efficiency}) assigns models per role rather than per step.

%%%%%%%%%%%%%%%%%%%%%%%%%%%%%%%%%%%%%%%%%%%%%%%%%%%%%%%%%%%%%%%%%%%%%%%%%%%%%%%
%%%%%%%%%%%%%%%%%%%%%%%%%%%%%%%%%%%%%%%%%%%%%%%%%%%%%%%%%%%%%%%%%%%%%%%%%%%%%%%
\section{Evaluation Protocol and Deviations}
\label{app:protocol}

\textbf{Harness.} Per-task limits: 30-minute wall-clock cap (a task that hits it counts as a failure) and 400 graph steps (LangGraph node steps, not device actions); each LLM call has a 180\,s timeout with 2 retries on the same model and no model fallback. AndroidWorld's own step budget (ten times the task's complexity) never binds in our harness, because the whole task runs inside one AndroidWorld step; the binding limits are the ones above. For the initial system, in \resStepsAExceedText{} successful runs (\resStepsAExceedShare\,\%), the agent used more device actions (taps, typing, swipes, back and key presses, app launches; notes and waits excluded) than that budget; the median was \resStepsAMedian{} actions per success. A cell gets up to 3 attempts for infrastructure failures only; an agent failure is never re-run. We report the timeout rate (\resTimeoutRate) and the distribution of time to success (\resTimeToSuccess). Infrastructure failures (emulator crash, harness error, outage of a model's pinned endpoint; serving conditions are fixed per model, Appendix~\ref{app:efficiency}) are not task failures: the task is re-queued, the event is logged, and the rate is reported. Each run logs tokens, cost and wall-clock time per task and per agent. Code: \texttt{google-research/android\_world} at commit \texttt{55ddaf5} (unmodified task and evaluator code), \texttt{android\_env} 1.2.3; task instances from the upstream suite seeding with seeds 30 and 31; a task counts as solved only if the agent ended the episode and the official success check returns a value above 0.5; partial scores below that count as failures. The non-root port of the cloud emulators is fixed for the whole final matrix (harness commit \ifanon withheld for review\else\texttt{db2792e}\fi). Parallelism: up to 27 cloud emulators at once (12 to 16 for most of the run), 6 phones and 1 stock emulator; the final runs took place from 2026-11-10 13:10 to 2026-11-11 15:33 UTC (2{,}264 cells). In the full-system runs on the 116 tasks, \resVideoErrText. Most ablation cells ran after the full-system cells; to check that run time does not bias the paired ablation deltas, \resLoadControlText. Under a rule declared before any re-run, \resOverloadRerunText; the paper's numbers use the re-runs.\ifversionB{} Recording the screen did not measurably slow the agent: over \resOvhCells{} paired runs of \resOvhTasks{} tasks (on/off/off/on), the median per-task change in agent time was \resOvhDwall\,\% (per agent step \resOvhDstep\,\%; tap latency \resOvhTap\,\%), with outcomes agreeing on \resOvhAgree{} tasks; the pre-set criteria (median per-task change below 5\,\%, median tool-latency change below 10\,\%) are \resOvhCriteriaMet. Per-task changes ranged from \resOvhDwallRange\,\% because repeated runs take different paths, so the check can rule out large effects only; text input (\resOvhInputText\,\% on \resOvhInputTextN{} calls) is too small a sample to conclude. The \resOvhVideos{} recordings have a median size of \resOvhVideoMB\,MB (range \resOvhVideoMBRange\,MB).\fi

\textbf{Confidence intervals.} Confidence intervals for success rates are 95\% percentile intervals from a two-stage bootstrap that resamples tasks, then runs within each task ($B = 10{,}000$). When every task has the same outcome, the bootstrap interval collapses to a point; we then report a 95\% Wilson score interval on tasks (successes = tasks solved, $n$ = tasks), because tasks are the resampling unit and the runs of one task are not independent. This applies only to the stock-emulator subset of the full system.

\textbf{Environments.} Each task runs in exactly one of three environments. \emph{Cloud emulators (90 tasks):} Android~14 (API~34 instead of the benchmark's API~33) x86\_64 emulator images without root, controlled with ADB over the network. The AndroidWorld task setup is ported to work without root: each non-system AndroidWorld app is re-signed with our key (code unchanged), and an instrumentation app signed with the same key runs shell commands with that app's user id, which gives access to the app's data for state snapshots and database reads and writes; the other workarounds are documented per task. The cloud emulators' clock cannot be set and they have no telephony and no Bluetooth. One cloud-emulator task, \texttt{ClockStopWatchPausedVerify}, cannot be passed on the cloud emulator image: its Clock app exposes a single element labelled `Stopwatch', whereas the official checker requires two, so every configuration scores 0 on it (0 of 24 runs); we keep it in the reported numbers as an environment limit. As in upstream AndroidWorld, tasks that clear device storage also delete OsmAnd's offline map; later OsmAnd tasks on the same device run without it unless the agent re-downloads it in-app. In our runs, \resOsmWipedText{} cloud OsmAnd results followed such a clear. In upstream AndroidWorld, the checker of RecipeDeleteMultipleRecipesWithConstraint matches the ingredient against lower-cased recipe directions, so for the \resRecipeCapsN{} of its \resRecipeIngredientsN{} ingredients that are capitalized it expects no deletion even when recipes use the ingredient; none of our initial-system  instances (\resRecipeAAffected{} of \resRecipeACells{} runs) is affected\ifversionB{}, and \resRecipeFinalText{}\fi. Two upstream calendar tasks that ask for the next upcoming event can draw a target time that has already passed on the device's date (for example, 12:00am, read as the start of the day), in which case the checker expects an event that is no longer upcoming; none of our initial-system  instances is affected (\resCalNowAAffected{} of \resCalNowACells{} runs)\ifversionB{}, and \resCalNowFinalText{}\fi. In upstream AndroidWorld, SimpleCalendarFirstEventAfterStartTime places the expected event exactly at the time the goal names (``first event after 3pm'' expects an event at 3pm), so it is solvable only under an inclusive reading of ``after''; this holds for every instance of the task (in our runs, \resFirstAfterASolved{} of \resFirstAfterACells{} runs across all initial-system  configurations were solved). In upstream AndroidWorld, OpenTracks tasks name some activity types with a space (``skate boarding'') while the app's other activities spell them without one (``skateboarding''), so a reader who counts both can disagree with the checker; none of our initial-system  instances is affected (\resSportsAAffected{} of \resSportsACells{} runs)\ifversionB{}, and \resSportsFinalText{}\fi. In upstream AndroidWorld, ExpenseAddMultipleFromMarkor marks the transactions to log by appending ``. Reimbursable.'' to their notes in the source file, while its checker expects the notes without that marker, so copying a note as written always fails; in our runs, \resExpMarkerASolved{} of \resExpMarkerACells{} runs across all initial-system  configurations solved this task.\ifversionB{} The final system's rule that selection markers in a source are not record content (Appendix~\ref{app:versionB}) addresses this case.\fi{} \emph{Physical phones (\nPhys{} tasks):} the date-relative tasks, which need a settable clock, and the Bluetooth tasks, which need a real Bluetooth radio, run on 6 Pixel~7 phones from \farm{} with Android~17 (API~37, user build, no root, no SIM), driven over ADB. They use the same non-root setup as the cloud emulators (re-signed apps and the app-uid bridge), and the device clock is set to AndroidWorld's fixed date with \texttt{cmd alarm set-time} before each task and checked again before scoring. One deviation remains: on the 5 Bluetooth tasks the Settings app's UI state is not reset from a snapshot, since that reset needs root. \emph{Stock emulator (\nEmu{} tasks):} the SMS tasks need telephony, so they run on the stock AndroidWorld emulator (official Android Emulator 37.2.13, Pixel~6 AVD, Android~13 / API~33 Google APIs image), whose modem is emulated; so do the two camera tasks, whose first-party app reset needs root, and ContactsNewContactDraft, whose evaluator reads accessibility hint text that the non-root port cannot supply. The app setup is re-run after every emulator boot. In upstream AndroidWorld, the setup step that creates contacts taps Save after a fixed one-second delay, which can race the sliding contact form; in all \resContactSetupN{} runs of the two affected tasks (SimpleSmsSendReceivedAddress and SimpleSmsResend), the agent's logs show the contacts their outcome depended on.

\textbf{Deviations from the official setup.} Task and evaluator code is upstream; only the transport to the device changes, and every workaround that fires is counted per task.
\emph{Cloud emulators (non-root port):}
\begin{itemize}\setlength\itemsep{0pt}
    \item[W1] Root shim: \texttt{adb root} and \texttt{whoami} report root, so upstream takes its root code paths, which W2 serves (all tasks, transport only).
    \item[W2] App-uid bridge: shell, pull and push on an app's private data run inside that app's own process through an instrumentation APK signed with the same key (tasks that seed or read app databases).
    \item[W3] Re-signing: non-system task APKs are re-signed with our key before install, code unchanged, so W2 can instrument them (same tasks as W2).
    \item[W4] Snapshot location: app snapshots are stored under \texttt{/data/local/tmp} instead of the app-data area, which is not writable without root.
    \item[W5] SELinux: \texttt{restorecon} and \texttt{chcon} are no-ops (SELinux is disabled on the image; no functional effect).
    \item[W6] Install flag: \texttt{adb install} adds \texttt{--bypass-low-target-sdk-block}, since Android~14 refuses APKs with targetSdk below 23 (the Clipper app).
    \item[W7] \texttt{monkey} gets \texttt{--pct-syskeys 0}, since the image declares no hardware keys (app launches through \texttt{monkey}).
    \item[W8] Incoming SMS is inserted into the SMS content provider instead of through the emulator console (SMS-receiving tasks).
    \item[W9] Clearing the SMS store goes through the telephony provider instead of on-device \texttt{sqlite3} (setup and teardown of the SMS tasks).
    \item[W10] Clock and Files map to the AOSP DeskClock and DocumentsUI, the only variants on the image (Clock and Files tasks).
    \item[W11] Setting the date is a no-op: the device keeps the real date, so the 11 date-relative tasks run elsewhere; the other tasks do not depend on the date.
    \item[W12] System apps (Chrome, Contacts, Dialer, Clock, Files, Settings) cannot be re-signed, so their private snapshot is not restored (upstream's own fallback when a snapshot is missing); their UI state is not reset between tasks, while their data, in providers or on shared storage, is reset by the tasks themselves.
    \item[W13] Upstream's root wipe of shared storage also removes app-private files under \texttt{Android/data}, through W2 (task resets).
    \item[W14] Information-retrieval goals keep the absolute date wording (``October 16 2023'') instead of upstream's random relative wording (``tomorrow''), which would be wrong on a real-date clock; this changes the goal wording of 12 tasks, not their ground truth.
    \item[W15] The one evaluator that reads the accessibility-forwarder tree gets a \texttt{uiautomator} dump instead; it lacks hint text, so ContactsNewContactDraft runs on the stock emulator.
    \item[W16] Paths under \texttt{/data/media/0} (root-only) are rewritten to \texttt{/storage/emulated/0} (same files).
    \item[W17] Shared storage is case-sensitive on these emulators (stock AVDs fold case), so the Markor data directory points at the app's real \texttt{Documents/markor} (Markor tasks).
\end{itemize}
\emph{Phones:} the same non-root layer (W1--W4, W12 for the Settings app on the 5 Bluetooth tasks, whose Bluetooth and Wi-Fi state is set by each task's own initialization), plus P5: ADB timeout floors of 90\,s for bridge calls and 60\,s for shell calls, because the phones are reached over a remote link (it only lengthens timeouts and never retries).
\emph{Stock emulator:} E1, an ADB timeout floor of 60\,s on every call, because under heavy load on the shared host the emulator's content providers answered in 20--40\,s against upstream's 10\,s timeouts (it only lengthens timeouts and never retries). E2: the UI-automation server used by the agent empties the accessibility tree that the ContactsNewContactDraft evaluator reads, so that task scored 0 regardless of the agent's actions; all of its runs were re-run with a harness that refreshes the environment before scoring (every run of that task, whatever its outcome).\ifversionB{} E4: right after task setup and before the agent's first action, one query checks that the contacts the task's setup should have created exist; if one is missing, the run is an infrastructure failure (setup incomplete) and is re-run, independently of outcome. To make that setup step reliable, contacts are created with the setup function's own longer delay option (3\,s instead of 1\,s), and a contact still missing after the check is created once more before the run is declared an infrastructure failure.\fi

\textbf{MobileMiniWoB++ deviations from the upstream harness.}
\begin{itemize}
    \item[D1] The device time is not initialized: the cloud emulators cannot set the clock. MiniWoB problems and rewards do not read the Android clock.
    \item[D2] The environment uses the ADB transport only (no emulator gRPC or console); MiniWoB only needs activity starts and broadcasts.
    \item[D3] Termination is polled every 2\,s of wall-clock instead of after each agent step, so a solved task stops early.
    \item[D4] Step budget: upstream allows 30 agent steps; \sys{} runs with the same limits as on AndroidWorld (400 steps, 1800\,s). Steps are not comparable across agents.
    \item[D5] A non-integer terminal reward (6 continuous-score tasks, none in the 62-task subset) is scored as a failure, since it cannot satisfy the official \texttt{reward == 1.0}; upstream records NaN. It fired on \resMiniNonIntRewardText.
    \item[D6] Fixed seeds: upstream MiniWoB instances are not seedable, so each seed installs a rebuilt app whose only change seeds the page's random generator (\texttt{Math.seedrandom}); the reward code is untouched, and the same seed gives the same instance for every configuration.
    \item[D7] An episode destroyed by the agent (the reward broadcast gets no answer while ADB is responsive, and the app process was replaced or the agent issued an app-leaving action) is scored as a failure; upstream raises. It fired on \resMiniDestroyedText.
\end{itemize}
\resMiniInfraLostText.

\begin{table}[h]
\centering
\caption{MobileMiniWoB++ (cloud emulators, Android~14, 2 seeds per configuration). SR: mean with 95\% CI (bootstrap over tasks and runs); cost and median wall-clock per task on all 92 tasks.}
\label{tab:mini}
\small
\fitcol{%
\begin{tabular}{lcccc}
\toprule
Configuration & SR, 62 tasks & SR, 92 tasks & \$/task & Median s \\
\midrule
Full system & \resMiniSRcellFull & \resMiniAllSRcellFull & \resMiniAllCostFull & \resMiniAllWallMedFull \\
w/o decomposition & \resMiniSRcellDecompOnly & \resMiniAllSRcellDecompOnly & \resMiniAllCostDecompOnly & \resMiniAllWallMedDecompOnly \\
Single-agent baseline & \resMiniSRcellSingleAgent & \resMiniAllSRcellSingleAgent & \resMiniAllCostSingleAgent & \resMiniAllWallMedSingleAgent \\
\bottomrule
\end{tabular}%
}
\end{table}

\textbf{Runs.} Every run is in the experiment registry; Table~\ref{tab:runs} lists them per configuration and seed.

\begin{table*}[h]
\centering
\caption{Runs per configuration and seed (solved tasks, success rate, infrastructure failures re-queued).}
\label{tab:runs}
\resTabRuns
\end{table*}

\textbf{Per-task flip tables.} Table~\ref{tab:flip} lists the tasks whose outcome differs between the two runs of the initial system's full configuration, and Table~\ref{tab:flipB} those solved in some but not all of the 3 runs of the final system.

\begin{table}[h]
\centering
\caption{Initial system: tasks solved in exactly one of the two runs (full configuration, 116 tasks, seeds 30 and 31; \checkmark{} solved, $\times$ failed).}
\label{tab:flip}
\fitcol{\resTabFlip}
\end{table}

\begin{table}[h]
\centering
\caption{Final system: tasks solved in some but not all of the 3 runs (116 tasks, held-out seeds \resBFinalSeeds{}; \checkmark{} solved, $\times$ failed).}
\label{tab:flipB}
\fitcol{\resBTabFlip}
\end{table}

\begin{figure}[h]
\centering
\fitcol{\resFigTTS}
\caption{Time to success of solved tasks, full configuration of the initial system (30-minute cap per task).}
\label{fig:tts}
\end{figure}

\begin{figure}[h]
\centering
\fitcol{\resFigHeatmap}
\caption{Success rate per application, full configuration of the initial system.}
\label{fig:apps}
\end{figure}

\section{Routing and Efficiency Details}
\label{app:efficiency}

This appendix provides detailed methodology and data for the routing and
efficiency analysis of Section~\ref{sec:efficiency}.

\subsection{Experimental Methodology}

Table~\ref{tab:routing} lists the configurations over the seven LLM roles (five graph agents, Hopper, Outputter; models per role in Appendix~\ref{app:efficiency}), each run with two seeds. Downgrading every non-Cortex agent at once tests the original claim that only the Cortex needs a frontier model. We log tokens and the cost reported by the API gateway per agent per task (price date: \resPriceDate) and how often each agent is invoked, since some (e.g., the Contextor's LLM) rarely run.

Every configuration runs on the 90 cloud-emulator tasks, paired with the full system on the same tasks and seeds, twice, under the protocol of Section~\ref{sec:protocol}. Every model is called through an API gateway, OpenRouter (accessed November 2026), and the model identifiers in Table~\ref{tab:pins} are its catalog ids; the original model ids are used where they are still served. Two are not: \texttt{google/gemini-3-pro-preview} (Cortex, and every role in All Frontier) is replaced by \texttt{google/gemini-3.1-pro-preview}, and \texttt{meta-llama/llama-3.1-70b-instruct} (Executor), whose only remaining endpoint has no tool calling, by \texttt{meta-llama/llama-3.3-70b-instruct}. Serving conditions are fixed per model (Table~\ref{tab:pins}); an outage of a model's endpoint ends the task as an infrastructure failure, which is re-queued and counted separately. The served model and endpoint of every call are logged.

\begin{table}[h]
\centering
\caption{Serving conditions per model. All runs used strict, fixed serving conditions: each open-weight model was served at a fixed precision (bf16/fp16 unquantized, or fp8 where no unquantized endpoint supports the required features) by one pinned inference endpoint, with provider fallback disabled. The one exception is \texttt{gpt-oss-120b}$^*$, which had two endpoints with the same precision, ordered, and fallback restricted to them (\resOssFallbackShare{} of its calls were served by the second endpoint). Proprietary models were called on their vendor's API. Features used: what the model's roles send (tool calling; structured output as a JSON schema). The endpoint that served each call is logged, and \resServedOutsideText.}
\label{tab:pins}
\small
\fitcol{%
\begin{tabular}{llll}
\toprule
Model & Role(s) & Precision & Features used \\
\midrule
\texttt{google/gemini-3.1-pro-preview} & Cortex; every role in All Frontier ($F$) & proprietary API & tools, structured output \\
\texttt{google/gemini-3-flash-preview} & video analyzer & proprietary API & neither (plain completion) \\
\texttt{meta-llama/llama-4-scout} & Planner & fp8 & structured output \\
\texttt{openai/gpt-oss-120b}$^*$ & Orchestrator & bf16 & structured output \\
\texttt{meta-llama/llama-3.3-70b-instruct} & Executor & fp16 & tools \\
\texttt{meta-llama/llama-3.1-8b-instruct} & Contextor & bf16 & structured output \\
\texttt{openai/gpt-5-nano} & Hopper, Outputter & proprietary API & structured output \\
\texttt{qwen/qwen3-vl-8b-instruct} & every downgraded role ($B$) & bf16 & tools, structured output \\
\bottomrule
\end{tabular}%
}
\end{table} Per task and per agent we log tokens (input, output, cached), number of LLM calls, call latency, and wall-clock time. Cost is the cost reported by the API gateway, logged per call by the runs. Table~\ref{tab:devtime} reports the cloud device time per configuration.

\begin{table*}[h]
\centering
\caption{Cloud device time per configuration.}
\label{tab:devtime}
\fitcol{\resTabDeviceTime}

\resTabDeviceTimeNote
\end{table*}

\subsection{Model Pricing}

Costs are the API gateway costs logged by the runs, with the price snapshot date stated (\resPriceDate). Table~\ref{tab:prices} gives the per-model prices recovered from the logged per-call costs.

\begin{table*}[h]
\centering
\caption{Per-model API gateway prices recovered from the logged per-call costs, and number of calls.}
\label{tab:prices}
\resTabPrices
\end{table*}

\begin{figure*}[h]
\centering
\fitcol{\resFigRoles}
\caption{Cost, latency and LLM calls per task by agent role, full configuration of the initial system.}
\label{fig:roles}
\end{figure*}

\subsection{Configuration Details}

Table~\ref{tab:configurations} gives the model of every LLM role in each routing configuration. $F$ is the single frontier model (\texttt{google/gemini-3.1-pro-preview}); $B$ is the single budget model used for downgrades (\texttt{qwen/qwen3-vl-8b-instruct}). The Platform Default profile was served by the now-decommissioned hosted platform; it is rebuilt from its documentation as a versioned configuration file of the agent repository (\texttt{benchmarks/configs/full.jsonc}), and the resolved configuration and its SHA-256 are logged in every run. The video analyzer uses \texttt{google/gemini-3-flash-preview} in every configuration.

\begin{table*}[t]
\centering
\caption{Model of each LLM role per routing configuration. Frontier model in \textbf{bold}. $^a$Llama 3.3 70B replaces the original Llama 3.1 70B, whose only remaining endpoint has no tool calling.}
\label{tab:configurations}
\small
\resizebox{\textwidth}{!}{%
\begin{tabular}{l|ccccccc}
\toprule
Configuration & Cortex & Planner & Executor & Orch. & Context. & Hopper & Output. \\
\midrule
All Frontier & $\mathbf{F}$ & $\mathbf{F}$ & $\mathbf{F}$ & $\mathbf{F}$ & $\mathbf{F}$ & $\mathbf{F}$ & $\mathbf{F}$ \\
Platform Default & $\mathbf{F}$ & Llama 4 Scout & Llama 3.3 70B$^a$ & GPT-OSS 120B & Llama 3.1 8B & GPT-5 Nano & GPT-5 Nano \\
Degrade Cortex & $B$ & Llama 4 Scout & Llama 3.3 70B$^a$ & GPT-OSS 120B & Llama 3.1 8B & GPT-5 Nano & GPT-5 Nano \\
Degrade all non-Cortex & $\mathbf{F}$ & $B$ & $B$ & $B$ & $B$ & $B$ & $B$ \\
\bottomrule
\end{tabular}%
}
\end{table*}

\subsection{Cost--Success Trade-off}

% v1 Figure 2 (02_agent_ablation.pdf, 50-task subset, Gemini 2.5 Pro) and the v1
% Pareto plot (pareto_frontier.pdf, mixed Gemini 2.5/3 Pro, single runs) are
% removed: they contradict each other (tracker R12). The Pareto plot will be
% regenerated from the P3 runs.
\begin{figure}[h]
\centering
\fitcol{\resFigPareto}
\caption{Cost--success trade-off across the routing configurations of Table~\ref{tab:routing} (90 cloud-emulator tasks); error bars: 95\% CI over runs. The Pareto frontier is drawn on means: it joins Degrade Cortex and Degrade all non-Cortex, while Platform Default and All Frontier lie off it (the success differences among the three upper points are not significant); budget configurations, Degrade Cortex: \resBudgetSeedLabelCortexBudget; Degrade all non-Cortex: \resBudgetSeedLabelNonCortexBudget.}
\label{fig:pareto_frontier}
\end{figure}

\subsection{Complete Results}

\begin{figure}[h]
\centering
\fitcol{\resFigAblation}
\caption{Leave-one-out deltas versus the full system on the 90 cloud-emulator tasks, with 95\% CIs (Table~\ref{tab:ablation}).}
\label{fig:ablation}
\end{figure}

\begin{table}[h]
\centering
\caption{Routing results on the AndroidWorld Cloud-90 subset (90 tasks), mean $\pm$ 95\% CI over two runs; budget rows, Degrade Cortex: \resBudgetSeedLabelCortexBudget; Degrade all non-Cortex: \resBudgetSeedLabelNonCortexBudget. $\Delta$SR relative to All Frontier. Calls: mean LLM calls per task of the downgraded role(s).}
\label{tab:results}
\small
\fitcol{%
\begin{tabular}{l|rrrr}
\toprule
Configuration & SR (\%) & $\Delta$SR & \$/task & Calls \\
\midrule
All Frontier & \resSRcellAllFrontier & --- & \resCostAllFrontier & --- \\
Platform Default & \resSRcellFull & \resRDeltacellFull & \resCostFull & \resCallsFull \\
Degrade Cortex & \resSRcellCortexBudget & \resRDeltacellCortexBudget & \resCostCortexBudget & \resCallsCortexBudget \\
Degrade all non-Cortex & \resSRcellNonCortexBudget & \resRDeltacellNonCortexBudget & \resCostNonCortexBudget & \resCallsNonCortexBudget \\
\bottomrule
\end{tabular}%
}
\end{table}

\section{Additional Discussion}
\label{app:discussion}

\textbf{Why decomposition may help.} Each agent keeps a focused context: the Planner plans without execution noise, the Cortex decides without planning traces, and the Executor acts without higher-level reasoning. \awdecomp{On the Cloud-90 tasks the data are consistent with this account: removing only the decomposition changes success by \resDeltaCIDecompOnly{} points; on the short single-screen tasks of MobileMiniWoB++ the decomposition brings no significant gain (Section~\ref{sec:second-benchmark}).}{The data do not support this account at significance level (Cloud-90: \resEffDecompositionCI{} points; MobileMiniWoB++: \resMiniEffDecompositionCI{}).}

\textbf{The verification principle.} The architecture separates intent from outcome: the LLM specifies what to do, and code reads the result back from the device (the content of a text field after typing, the foreground package under app lock), so the LLM adapts to the actual state instead of assuming success. On AndroidWorld, removing the text comparison alone does not change success significantly (\resDeltaCINoTextCheck{} points), so this is a design rationale rather than a measured effect; we expect it to matter for agents acting in stochastic environments.

\ifversionB
\section{Reliability Changes: Full List}
\label{app:versionB}

\textbf{Definition.} The final system uses the same models, agents and routing as the initial system, plus general changes, each stated as a general behaviour and motivated by a class of failures, never by a task. Every change sits behind a configuration flag that is off in every configuration of the initial system, whose prompts it leaves byte-identical. The wording below is that of the frozen version (\resBFreezePin{}).

\textbf{Architecture-adjacent changes} (listed so that readers can judge ``same architecture''):
\begin{itemize}\setlength\itemsep{0pt}
    \item Executor completion: predictable actions the Cortex decided but the Executor's LLM left out are dispatched in order, schema-validated, stopping after the first screen-changing action; an emitted action is never dispatched again.
    \item Goal checklist: the goal is turned once into explicit requirements (Cortex model, inside the Planner).
    \item Goal verifier: judges each requirement on device evidence only, rejects at most 3 times, and its final answer is the reported answer.
    \item Step-budget checkpoint: with fewer than 70 graph steps left, the verifier is asked once.
    \item Replan guard: a fresh plan gets at least one Cortex turn before another replan can be requested.
    \item Capture goals: the Planner plans no screen recording; recording is used only to read a video that is already playing.
\end{itemize}

\textbf{Deterministic context given to the agents.} Calendar facts from the device date (weeks Monday--Sunday and Sunday--Saturday, months); the sort settings in force (selected key, state of a reverse option), with an ``establish the order first'' subgoal for superlative goals; for the verifier, a log of the UI text of recent distinct screens and a log of executed actions; a capture that waits until the UI tree is stable; and, after 2 identical tool failures, a note that a change of strategy is required.

\textbf{New tools.} \texttt{read\_full\_list} reads a whole list or document from top to bottom and confirms its end; \texttt{inspect\_rows} opens each matching row's detail screen, reads it and goes back, grouping records identical on every field; \texttt{calculate} does arithmetic and unit or time conversion; \texttt{sample\_colors} reads exact colours from a lossless screenshot. All are read-only.

\textbf{Tool robustness.} Taps correct a centre given as a corner and fall back to the live element when the given bounds miss it; text input refuses targets that are not on screen and reports which field received the text; the video tool stops only its own recorder and removes its files; tool arguments accept the shapes models send; the soft keyboard is hidden before the final check. Trailing line breaks in single-line text are sent as Enter presses; multi-line text is typed whole, and if the readback has fewer trailing line breaks than typed, the missing ones are typed once more (at most once per field, logged).

\textbf{Planner and Cortex rules.} Read the whole list or source before answering, counting, choosing by a property or declaring a multi-item goal done; an item is absent only after its list was read to its end. Names, values and text are taken exactly as written in the goal. Exact dates take precedence over relative day labels. A category or type comes from the item's field, never from its title or icon. Order is never inferred from position. Duplicates are identical on every field. Entering data in a form includes submitting it, unless the goal asks for a draft. A revealed option is tapped and verified selected. Question goals are read-only. If an app reports that needed data is missing, its own download is used once; before a distance-ranked search, the view moves to the target region; a placed point is checked where it landed. Pre-existing records are left untouched in ``add'' goals; exactly the requested number of items is created, and items this run did not create are never deleted or edited. An item exactly at a goal's time, date or value bound is included unless the goal says ``strictly'' or ``exclusive''. In map or canvas create modes, empty space is never tapped to dismiss a card. The video-analysis tool is used only when the content of a playing video is needed, and no system side effect is started unless asked. Two rules are borderline-general: \label{app:versionB-markers}selection markers in a source are not record content, and ``a new line between blocks'' means an empty line.

\textbf{Verifier rules.} Evidence is what the device stores (screens, deterministic logs, tool outputs); agent statements are claims. A rejection needs a contradicted or unestablishable requirement and names one targeted check. Lists are judged on complete content, fields on the value of the field the goal names, positions by place, and a committed result must be visible. The action log proves that an action ran, not what it hit: positions, the record edited or deleted, and presence need a device screen after the last change. Boundaries follow the Planner's inclusive convention. A goal's own example fixes the granularity and format of the answer. Counts of created items come from the executed-actions log. An empty answer needs two different methods. Changes the goal did not ask for violate ``nothing extra''.

\textbf{Rules triggered by the goal's wording} (disclosed; other goals' prompts are unchanged):
\begin{itemize}\setlength\itemsep{0pt}
    \item Names given with a file extension: the saved name itself must end with it; a format label beside the name is not part of it.
    \item Records copied from a named source file: three fixed checklist lines are appended (the goal's record noun is not a filter, a selection marker is left out of copied fields, field values are complete as the source writes them). The first two are borderline-general.
    \item Date-window questions whose dates show only as relative labels: the app's own full-date display is switched on to read, then restored; this is a disclosed exception to ``questions are read-only''.
    \item Goals that create or save a track or route through given places: those places must be points of the track's own path, in order, and a fixed procedure is followed in route planners that add a point at the map centre (borderline-general; triggered in code by the goal's wording).
    \item Verifier gates: for goals deleting named items, a satisfied verdict with no executed delete becomes one rejection asking for a complete read of the list (deterministic, once per run, never a loop); for source-copy goals in which a delete ran, one rejection asks to re-check pre-existing and own records (soft, once per run).
\end{itemize}

\textbf{Known limitations.} In the route procedure, the agent can tap the planner's add-point button while a place card still covers it, so a point can land away from the place and still be accepted by the verifier (observed in \resBRouteRiskText). \texttt{inspect\_rows} does not resume past rows it already reported, so a long list that moves after Back can be re-read until a limit is reached.

\textbf{Limits.} The final system raises the internal limits of the initial system: \resBLimitSteps{} graph steps (initial: 400) and a \resBLimitSeconds\,s wall-clock cap per task (initial: 1800\,s), with no cap on device actions. As for the initial system, AndroidWorld's own step budget never binds; for the final system, in \resStepsBExceedText{} successful runs (\resStepsBExceedShare\,\%), the agent used more device actions than that budget; the median was \resStepsBMedian{} actions per success.

\textbf{Development protocol.} The final system was developed over \resBNIterations{} iterations, with \resBNFreezeChecks{} freeze checks, on development seeds \resBDevSeeds, reused across iterations. Seeds 30 and 31 are those of the initial system's study, so the re-checks on them served as a development gate only, never as evidence on fresh seeds. The final seeds (\resBFinalSeeds) were never used before the freeze, which the run matrix enforces; the freeze record holds the final system's commit. \resBFreezeText{} \resBOsmTrackText{} Upstream instances that cannot measure correct behaviour (the recipe, calendar and OpenTracks notes of Appendix~\ref{app:protocol}) were excluded from the development gate only; the final runs count every instance as is.

\textbf{Environment.} The final runs use \resBNCloud{} tasks on cloud emulators, \resBNPhys{} on phones and \resBNEmu{} on the stock emulator: ClockStopWatchPausedVerify, OpenAppTaskEval and TurnOnWifiAndOpenApp move to the stock emulator, where deterministic probes show them passable, unlike on the cloud image (the open-app checks expect the Google Clock and a camera; the initial system's runs are unaffected). Deviation E2 applies to every stock-emulator run, and E4 to every run (Appendix~\ref{app:protocol}). A cell whose infrastructure attempts are exhausted is re-queued up to \resBRequeueRounds{} more rounds; if it still fails, it counts as a failure and is flagged unscored (\resBUnscoredN{} cells, \resBUnscoredShare\,\%); \resBInfraLimitedText.

\begin{table}[h]
\centering
\caption{Final system on the held-out seeds \resBFinalSeeds{} (\resBK{} runs per task): pass@\resBK{}, pass\^{}\resBK{} and mean success rate with 95\% CI, and the number of tasks never solved.}
\label{tab:versionB}
\small
\fitcol{%
\begin{tabular}{lcccc}
\toprule
Tasks & pass@\resBK{} & pass\^{}\resBK{} & Mean SR & Never solved \\
\midrule
AndroidWorld, 116 & \resBAWPassAtKtext & \resBAWPassHatKtext & \resBAWSRtext & \resBAWNeverSolved \\
Cloud emulators, \resBNCloud{} & \resBCloudPassAtKtext & \resBCloudPassHatKtext & \resBCloudSRtext & \resBCloudNeverSolved \\
Phones, \resBNPhys{} & \resBPhysPassAtKtext & \resBPhysPassHatKtext & \resBPhysSRtext & \resBPhysNeverSolved \\
Stock emulator, \resBNEmu{} & \resBEmuPassAtKtext & \resBEmuPassHatKtext & \resBEmuSRtext & \resBEmuNeverSolved \\
Fresh instances, 92 & \resBFreshPassAtKtext & \resBFreshPassHatKtext & \resBFreshSRtext & \resBFreshNeverSolved \\
\bottomrule
\end{tabular}%
}
\end{table}

\textbf{MobileMiniWoB++.} On the 92 tasks (seeds 30 and 31, paired with the initial system), the final system changes success by \resBMiniDeltaCI{} points ($p = \resBMiniDeltaP$; 62-task subset: \resBMiniSubsetDeltaCI{}): \resBMiniVerdictText.
\fi